\pdfoutput=1
\documentclass[twoside,11pt]{article}

\usepackage[preprint,nohyperref]{jmlr2e}

\let\proof\undefined

\usepackage[T1]{fontenc}%

\usepackage{amsmath,amssymb,amsfonts}%
\usepackage{mathtools}%
\usepackage{bm}%
\usepackage{amsthm}%
\usepackage{mathrsfs}%

\usepackage{graphicx}%
\usepackage{subcaption}%
\usepackage{booktabs}%
\usepackage{multirow}%
\usepackage{makecell}%
\usepackage{array}%
\usepackage{longtable}%
\usepackage{tabularx}%
\usepackage{threeparttable}%
\usepackage{colortbl}%
\usepackage{adjustbox}%
\usepackage{rotating}%
\usepackage{pdflscape}%
\usepackage{siunitx}%
\usepackage{pifont}%
\usepackage{float}%

\usepackage{placeins}%

\usepackage{algorithm}%
\usepackage[noend]{algpseudocode}%

\usepackage{xcolor}%
\usepackage{enumitem}%
\usepackage{microtype}%
\usepackage{textcomp}%

\usepackage{tikz}%
\usetikzlibrary{arrows.meta,positioning,fit,backgrounds,shapes.geometric}
\usepackage{pgfplots}\pgfplotsset{compat=1.18}

\usepackage{hyperref}%
\usepackage{cleveref}%
\usepackage{tntheme}%

\theoremstyle{definition}
\ifcsname remark\endcsname\else\fi
\ifcsname hypothesis\endcsname\else\fi



\definecolor{winrow}{HTML}{EAF2FB}
\definecolor{winnerfill}{HTML}{EAF2FB}

\newcommand{\tnfig}[2]{%
  \IfFileExists{figures/#1}%
    {\includegraphics[width=\textwidth]{figures/#1}}%
    {\fbox{\parbox[c][0.32\textheight][c]{0.98\linewidth}{\centering\itshape
       [figure pending: \detokenize{#1}]\\[4pt]#2}}}}
\newcommand{\srfig}[2]{%
  \IfFileExists{figures/#1}%
    {\includegraphics[width=\linewidth]{figures/#1}}%
    {\fbox{\parbox[c][0.30\textheight][c]{0.97\linewidth}{\centering\itshape
       [extended figure pending: \detokenize{#1}]\\[4pt]#2}}}}

\graphicspath{{figures/}{figs/}{./figures/}{./figs/}}

\newcommand{\numTasks}{1{,}024}                 %
\newcommand{\numGridded}{512}                   %

\newcommand{\numCountry}{512}                   %

\newcommand{\numTemporalTasks}{29}               %
\newcommand{\medianAnom}{0.74}                %
\newcommand{\shortRangeMedian}{0.79}            %
\newcommand{\anomEdge}{0.79}                       %
\newcommand{\anomBackcastFar}{0.60}             %
\newcommand{\anomForecastFar}{0.49}             %
\newcommand{\mapsLevelInterp}{0.999}            %
\newcommand{\mapsLevelExtrap}{0.98}             %
\newcommand{\mapsErrInterp}{0.6}                %
\newcommand{\mapsErrExtrap}{3\text{--}4}        %
\newcommand{\mapsWidthGrow}{26\%}               %

\newcommand{\sparseHeroFrac}{1/1024}            %
\newcommand{\sparseHeroPSNR}{23}                 %
\newcommand{\sparseHeroRtwo}{0.94}              %
\newcommand{\sparseRtwoDense}{0.97}             %
\newcommand{\sparseInterpDense}{0.95}           %
\newcommand{\sparseRtwoSparse}{0.74}            %
\newcommand{\sparseInterpSparse}{0.56}          %
\newcommand{\sigmaWiden}{0.16\text{--}0.45}     %

\newcommand{\adaptParams}{416{,}000}            %

\newcommand{\retrievalRecallOne}{0.875}         %
\newcommand{\retrievalRecallFive}{1.00}         %
\newcommand{\retrievalRecallTen}{1.00}
\newcommand{\retrievalPool}{80}                 %

\newcommand{\retrievalPoolAbl}{22}              %
\newcommand{\idTerranova}{9.8\text{--}9.9}      %

\newcommand{\benchMeanR}{0.88}                  %
\newcommand{\benchMeanRBase}{0.77}              %
\newcommand{\benchDeltaRange}{0.02\text{--}0.26}%
\newcommand{\benchOceanTN}{0.88}                %
\newcommand{\benchOceanBase}{0.62}              %
\newcommand{\benchOceanDrop}{0.12\text{--}0.60} %
\newcommand{\benchCrossover}{512}               %
\newcommand{\reconNumTasks}{31}                  %
\newcommand{\reconKdisp}{32}                     %
\newcommand{\reconTnCurve}{0.17\text{--}0.51}    %
\newcommand{\reconIdwCurve}{0.13\text{--}0.47}   %
\newcommand{\reconTnWins}{7}                     %
\newcommand{\reconDissoc}{-0.57}                 %
\newcommand{\reconErrCorrHigh}{0.81}             %
\newcommand{\reconErrCorrLow}{0.61}              %
\newcommand{\reconMeat}{0.65}                    %

\newcommand{\reconTour}{0.71}                    %
\newcommand{\reconTourBase}{0.22}

\newcommand{\downQuartet}{0.51\text{--}0.66}    %
\newcommand{\downNullMax}{0.10}                 %
\newcommand{\downPositiveShare}{91\text{--}96\%}%
\newcommand{\downCountries}{149\text{--}154}    %
\newcommand{\downMedianWithin}{0.30\text{--}0.40}%
\newcommand{\downExchange}{55}                  %
\newcommand{\devHDI}{0.95}                     %
\newcommand{\devGDPpc}{0.90}                   %
\newcommand{\devEvr}{54\%}                     %
\newcommand{\devPanel}{nine}                   %
\newcommand{\devHDInoLife}{0.95}               %
\newcommand{\devHDInoLongevity}{0.91}          %
\newcommand{\devRawEvr}{5\%}                   %
\newcommand{\devRawHDI}{0.23}                  %

\newcommand{\trainGpuModel}{H100 PCIe}          %
\newcommand{\laptopGpu}{RTX 5070 Laptop}        %
\newcommand{\trainHours}{327}                   %

\newcommand{\trainKWh}{128}                     %
\newcommand{\trainCarbon}{42}                   %
\newcommand{\trainEpochs}{128}
\newcommand{\trainSteps}{4.23}                  %
\newcommand{\trainSamples}{34}                  %
\newcommand{\trainStepsPerSec}{3.6}             %

\newcommand{\modelParams}{366}                  %
\newcommand{\flopsCoord}{713}                   %
\newcommand{\flopsCountry}{625}                 %
\newcommand{\flopsFusionShare}{57\%}            %

\ShortHeadings{TerraNova: A Foundation Model for the Anthropocene}{Rodriguez-Pardo and Tavoni}
\firstpageno{1}

\newcommand{\SIfootnotemark}{\footnote{The Supplementary Information --- the
full methodological specification, the complete ablation programme, extended
results and the computational-cost analysis --- is available at the project
page: \url{https://carlosrodriguezpardo.es/projects/TerraNova/}.}\global\let\SIfootnotemark\relax}
\newcommand{\SIplace}{the Supplementary Information\SIfootnotemark}

\begin{document}

\title{\emph{TerraNova}: A Foundation Model for the Anthropocene}

\author{\name Carlos Rodriguez-Pardo \email carlos.rodriguezpardo.jimenez@gmail.com \\
       \addr Department of Management, Economics and Industrial Engineering, Politecnico di Milano\\
       RFF-CMCC European Institute on Economics and the Environment (EIEE)\\
       Euro-Mediterranean Center on Climate Change (CMCC)
       \AND
       \name Massimo Tavoni \email massimo.tavoni@polimi.it \\
       \addr Department of Management, Economics and Industrial Engineering, Politecnico di Milano\\
       RFF-CMCC European Institute on Economics and the Environment (EIEE)\\
       Euro-Mediterranean Center on Climate Change (CMCC)}

\maketitle

\begin{abstract}%
A defining problem of the Anthropocene is to model the physical Earth and human societies as one coupled system, yet no learned representation spans their observational breadth. We argue the obstacle is geometric: the physical Earth is measured as continuous fields that ignore political borders, whereas societies are reported for administrative units. Earth-system foundation models serve the first geometry; coupling it to the second has required lossy averaging over borders. We introduce \emph{TerraNova}, a foundation model trained on 1,024 physical and societal records in their native geometries: 512 gridded Earth-system fields and 512 national indicators. Dedicated encoders represent location, country, time and task, cross-modal transformers fuse them into a shared spatiotemporal state, and a hypernetwork generates a per-query decoder whose evidential head returns a predictive distribution. Two contrastive objectives couple the representation: a population-weighted alignment between each country and coordinates in its territory, and one to pretrained geospatial embeddings carrying image-derived semantics. Read out through that decoder, the representation is competitive with purpose-built geospatial encoders while spanning axes they do not represent (time, oceans and uncertainty) and supporting country-level capabilities. The frozen backbone reconstructs dense fields from sparse observations and adapts to unseen variables in minutes on consumer hardware.

\end{abstract}


\vskip 0.1in
\noindent\textbf{Project page:}~\url{https://carlosrodriguezpardo.es/projects/TerraNova/}\hfill\break
\noindent\textit{The Supplementary Information (methods, ablations, extended results, cost) is hosted there.}

\section{Introduction}\label{sec:intro}

A key scientific problem of the Anthropocene~\citep{crutzen2002geology} is to jointly model societies and the Earth system as the interconnected whole they form: greenhouse-gas emissions alter atmospheric composition; land transformation changes hydrology, albedo and carbon storage; infrastructure concentrates exposure; institutions shape vulnerability; and climate impacts affect economic development, migration and policy~\citep{steffen2015planetary, richardson2023earth}. These connections are already visible in observations: climate change has altered the distribution of economic growth and inequality across countries~\citep{diffenbaugh2019global}, its effects are already visible in human development outcomes~\citep{tavoni2025past}, and estimates of its economic cost are large and widely dispersed~\citep{moore_synthesis_2024}. Planetary change and policy-relevant research requires data representations that span physical, ecological, economic and institutional systems, rather than treating their records as unrelated datasets~\citep{ou2026ai}.

Existing model families cover only a part of that system. Weather and climate models resolve physical dynamics through numerical approximations to physical law, and are compared through coordinated multi-model protocols~\citep{eyring2016overview}; economic and integrated-assessment models represent welfare and policy through regional aggregates, scenarios and damage relationships~\citep{nordhaus2017dice, riahi2017shared}. Earth-system foundation models have improved the accuracy and speed of geophysical prediction, from global forecasting~\citep{lam2023graphcast, bi2023pangu} to multi-variable and multi-domain adaptation~\citep{nguyen2023climax, bodnar2025foundation}, but they remain confined to the physical geometry. No single learned representation spans the physical Earth and the societies embedded in it, and to our knowledge none learns from both simultaneously at a global scale.

The physical Earth is measured as a \emph{field}: climate, vegetation, or soil moisture vary continuously over space and time. Societies are observed within \emph{boundaries}: their governance, health and development are reported yearly for countries and other units whose borders are political rather than physical. Italy is not a coordinate, and a grid cell in the Alps is not an economy, yet local environmental conditions and national institutions describe the same places at different scales. Standard processing resolves the mismatch by averaging fields over countries or rasterising administrative statistics, but both transformations are lossy: aggregation hides within-country gradients, whereas rasterisation implies precision that was never observed. We therefore formulate coupled environmental--societal modelling as a \emph{multi-geometry representation-learning} problem (Figure~\ref{fig:overview}).

Here we present \emph{TerraNova}, a foundation model trained on \numTasks{} variables relevant to the Anthropocene, from physical Earth-system fields to national socioeconomic and institutional indicators, each in its native geometry. Purpose-built encoders represent location, country, time and task, cross-modal transformers fuse them into a shared spatiotemporal state, and a hypernetwork generates a per-query decoder whose evidential head returns a predictive distribution with aleatoric and epistemic uncertainty proxies. Two contrastive objectives couple the geometries and modalities: a population-weighted, time-aware objective aligns each country with coordinates sampled from its territory, establishing the field--boundary correspondence, while a second aligns TerraNova's location representations with pretrained geospatial embeddings, importing geographic semantics learned from visual information. Conceptually, TerraNova treats physical fields, administrative records and geospatial imagery as different observations of one latent reality. The shared statistical structure it learns supports transfer, spatial inference and hypothesis generation rather than causal or scenario analysis. We make the following contributions:

\begin{figure}[t]
	\centering
	\includegraphics[width=0.9\linewidth]{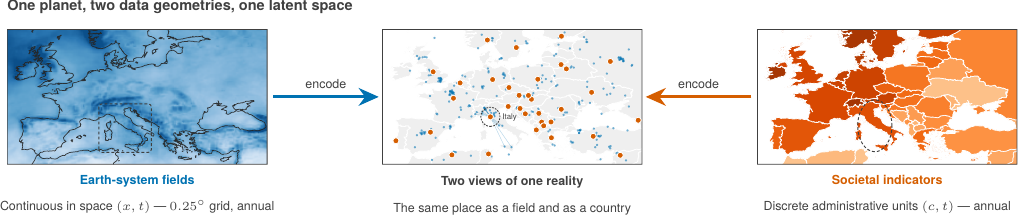}
	\caption{The physical Earth is observed as continuous fields, whereas societal indicators are reported for discrete administrative units. Both are observations of one shared latent reality.}
	\label{fig:overview}
    \vspace{-5mm}
\end{figure}

\begin{itemize}[leftmargin=1.2em,itemsep=1.8pt,topsep=1.8pt]
\item To our knowledge, TerraNova is the first model to learn one representation across the observational breadth of the Anthropocene, trained on \numTasks{} variables in two geometries within a single backbone.
\item That breadth is made possible by inter-geometry coupling. A country--location contrastive objective connects administrative units to their territories, while alignment to geospatial embeddings anchors the representation in image-derived semantics. The coupling supports capabilities neither geometry expresses alone, from country--territory retrieval to national-to-gridded downscaling.
\item Read out through its task-conditioned decoder, the shared representation leads purpose-built geospatial embeddings on most static targets and label budgets, and extends to axes they do not represent: uncertainty, time and the oceans.
\item Every query returns a predictive distribution, a hypernetwork-generated evidential head supplying aleatoric and epistemic uncertainty proxies alongside each prediction. Adaptation to new variables can be reached cheaply through lightweight adapters on a frozen backbone.
\end{itemize}

\section{Related Work}\label{sec:related}
\paragraph{Foundation models for the physical Earth system.} Data-driven models now rival operational numerical weather prediction at a fraction of its inference cost. GraphCast learns medium-range forecasting on an icosahedral mesh~\citep{lam2023graphcast}; Pangu-Weather applies a three-dimensional transformer over pressure levels~\citep{bi2023pangu}; and FourCastNet applies Fourier neural operators to forecasting~\citep{pathak2022fourcastnet}. ClimaX tokenises heterogeneous climate variables for a shared backbone~\citep{nguyen2023climax}, while Aurora adapts one pretrained model across weather, air quality, ocean waves and cyclones~\citep{bodnar2025foundation}. GenCast adds probabilistic ensemble forecasting~\citep{price2025probabilistic}, and Aardvark learns an end-to-end observation-to-forecast pipeline~\citep{vaughan2025endtoend}. These systems are designed for geophysical fields on grids or meshes. TerraNova does not aim to outperform them; instead it addresses a different and complementary problem.

\paragraph{Geospatial embeddings.} Geospatial embedding foundation models learn representations of location from Earth observation data~\citep{klemmer2025earth}. SatCLIP aligns coordinates with satellite imagery~\citep{klemmer2025satclip}, GeoCLIP with ground-level imagery~\citep{cepeda2024geoclip}; Clay~\citep{jakubik2024clay} learns multi-sensor Earth-observation features and Prithvi-EO-2.0 adds multi-temporal modelling with explicit location and time embeddings~\citep{szwarcman2024prithvi}; Copernicus-FM unifies Sentinel sensors through dynamic hypernetworks, including a global $0.25^{\circ}$ embedding product~\citep{wang2025copernicus}. TerraMind extends this family to any-to-any generation across nine Earth-observation modalities~\cite{jakubik2025terramind} and AlphaEarth learns an embedding field, released as global annual layers~\cite{brown2025alphaearth}. Related to our work, the Population Dynamics Foundation Model combines environmental signals such as weather and air quality with aggregated human-activity signals, only over the United States~\citep{agarwal2024population}. Location encoders based on spherical harmonics provide resolution-free coordinate features~\citep{russwurm2024geographic}. Existing models share three limitations: trained largely on imagery, they carry a weak signal over water, which covers most of the planet; they do not encode time; and they do not model uncertainty. We leverage this literature in two ways. As the closest comparison class for TerraNova's continuous geographic representation, these encoders are our evaluation baselines; and, as pretrained embeddings, we use them as alignment targets during training, anchoring TerraNova's location representation in semantics learned from imagery. TerraNova is thus time-dependent, uncertainty-aware, and defined at every location on the planet, and it connects these geospatial embeddings' view of place to environmental fields and country variables in one model.

\paragraph{Climate--economics integration.} Integrated assessment models (IAMs) couple reduced-form climate and economic systems through scenarios, technologies and damage functions~\citep{nordhaus2017dice, barrage2024policies, riahi2017shared, oneill2014new, weyant_contributions_2017}; richer systems add detailed energy and land-use sectors~\citep{stehfest2014image}, and the family has drawn a substantial critical literature of its own~\citep{gambhir2019review}. Empirical climate economics instead estimates relationships from observations, including nonlinear temperature--growth effects~\citep{burke2015global, hsiang2017estimating}, climate-driven inequality~\citep{diffenbaugh2019global}, and impacts on human development~\citep{tavoni2025past}. Closest to our output geometry, gridded socioeconomic products disaggregate national statistics such as GDP and HDI onto global grids~\citep{kummu2018gridded}; these are valuable per-variable constructions, each built from a fixed covariate recipe, without a shared representation, any learned temporal model beyond the years tabulated, or per-cell uncertainty. TerraNova is not an alternative to the causal, normative or scenario-based frameworks above: instead, it provides a joint observational layer from which downstream empirical and policy models can draw, without assigning causal meaning to the learned associations.

\paragraph{Multi-modal representation learning.} A rich literature aligns representations across modalities, notably between language and vision: CLIP aligns images and text~\citep{radford2021clip}; ImageBind connects several modalities through a shared anchor~\citep{girdhar2023imagebind}; and Perceiver~IO maps heterogeneous inputs through a common array~\citep{jaegle2022perceiver}. The \emph{Platonic Representation Hypothesis} argues that neural networks trained on different data and modalities are converging towards a shared statistical model of reality~\citep{huh2024platonic}. TerraNova instantiates a vision of this idea for the Earth: environmental fields, administrative records and image-derived embeddings describe the same world, but differ in content, geometry, resolution, and availability. Rather than expecting compatibility to emerge from data, TerraNova couples these modalities through highly multimodal training (\numTasks{} reconstruction tasks) and explicit auxiliary learning objectives.

\paragraph{Predictive uncertainty.} Uncertainty quantification is key for making robust decisions. Two sources should be distinguished: \emph{epistemic} uncertainty, which reflects limited knowledge and might be possibly reduced with more data, and \emph{aleatoric} uncertainty, the irreducible randomness of a known data-generating process; the two carry different implications for whether to gather observations or to hedge against variability~\citep{tavoni_uncertainty_2022}. Evidential deep learning places a higher-order prior over a Gaussian likelihood and produces a predictive distribution in a single forward pass~\citep{amini2020evidential}, although subsequent work identifies regularisation and identifiability limitations~\citep{meinert2023unreasonable}. Earth-system science and integrated assessment quantify uncertainty through ensembles, parameter uncertainty, internal variability and alternative scenarios~\citep{hawkins2009potential,hawkins2011potential,chiani2025global}, and communicate the result through calibrated confidence and likelihood language~\citep{mastrandrea2010ipcc}. TerraNova targets task-conditional, per-query predictive uncertainty for heterogeneous variables: following \citet{amini2020evidential}, it outputs the parameters of a Normal--Inverse--Gamma distribution, providing a total predictive distribution alongside aleatoric and epistemic uncertainty proxies.

\section{Model design}\label{sec:model}

In this section, we introduce each component of TerraNova's design; further detail is provided in \SIplace. Two principles run through the whole architecture: every encoder output is rescaled to a sphere of fixed radius, and every residual pathway is initialised as a near-identity transformation.

\subsection{Model definition}
\begin{figure}[t]
\centering
\includegraphics[width=0.9\linewidth]{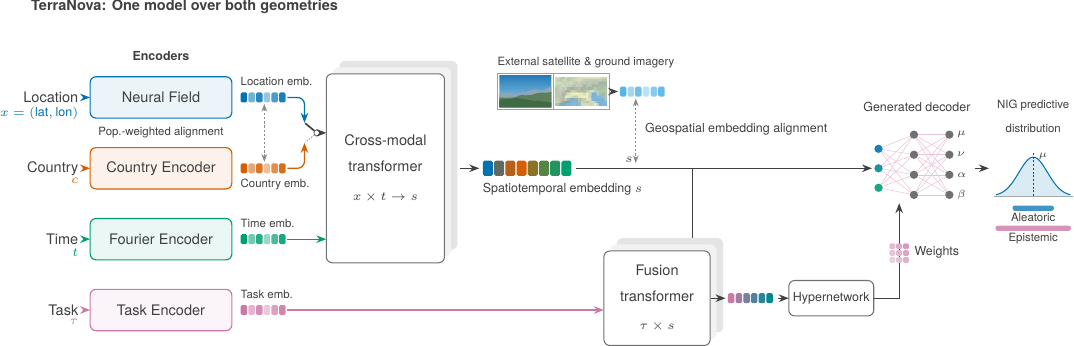}
\caption{\textbf{TerraNova architecture.} Dedicated encoders for location, country, time and task feed geometry routing and temporal fusion; a task-conditioned hypernetwork then generates the local decoder, whose NIG head returns a predictive distribution.}
\label{fig:arch}
\end{figure}

TerraNova, illustrated in Figure~\ref{fig:arch}, is a neural mapping over tasks, time and space,
\begin{equation}
  f_\theta \colon (\tau, u, t) \;\longmapsto\; (\mu, \nu, \alpha, \beta),
  \label{eq:signature}
\end{equation}
in which $\tau$ indexes the prediction task, $t$ is the observation year, and $u$ is the geometry-dependent spatial input: a coordinate $u = x = (\lambda, \phi)$ represents a gridded field, while an administrative unit $u = c$ addresses an indicator reported at the national level. The four outputs parameterise a Normal--Inverse--Gamma predictive distribution over the standardised target. We assume the two data types share a statistical structure that can be decoded into accurate predictions, but we do not enforce a shared geometry: TerraNova preserves each native query geometry, routes both into a common spatiotemporal state, and couples the geometries through soft alignment objectives.

All encoder outputs are radius-normalised,
\begin{equation}
  \mathrm{N}_d(\mathbf{v}) = \sqrt{d}\, \frac{\mathbf{v}}{\lVert\mathbf{v}\rVert_2},
  \label{eq:radius}
\end{equation}
which keeps cosine similarities, residual gates and attention logits at a stable scale across encoders of very different internal magnitude, and prevents any modality from dominating the fused representation by norm alone. Embedding widths are deliberately small relative to what they encode (256 dimensions for location against a global $0.25^{\circ}$ grid, 256 for task against 1{,}024 target variables, 64 for country, 32 for time, with a 256-dimensional shared spatiotemporal state feeding a 512-dimensional conditioning vector) so the model operates as an encoder--decoder that compresses the Earth system, enabling parameter efficiency and cheap transfer to unseen tasks (\S\ref{sec:adaptation}).

\subsection{Encoders}
Four purpose-built encoders map location, country, time and task to unit-radius embeddings (Eq.~\ref{eq:radius}).

\paragraph{Location.} The location encoder serves two regimes at once: smooth fields (temperature, sea-level pressure) whose spatial spectrum concentrates at low frequencies, and sparse ones (population density, built surface, burned area) with heavy high-frequency content and sharp discontinuities at coastlines and city edges. A network biased towards smoothness would blur the latter; one with enough local capacity for the latter injects noise into the former. We therefore run two complementary encodings of the same coordinate in parallel and let the encoder route between them per latent dimension. A continuous branch (a residual U-Net implicit neural network~\citep{rodriguez2023neubtf,sitzmann2020siren} with Gabor-wavelet activations~\citep{saragadam2023wire}) provides a smoothness-biased prior; a multi-resolution hash grid~\citep{muller2022instant} provides local high-frequency capacity. Then, a learned per-dimension gate blends them,
\begin{equation}
  \mathbf{e}_\ell = \mathrm{N}_{256}\!\left(W_{\mathrm{comb}}\bigl(
  \mathbf{g} \odot \mathbf{h}_{\mathrm{wire}} + (\mathbf{1}-\mathbf{g})
  \odot \mathbf{h}_{\mathrm{hash}}\bigr)\right),
  \label{eq:gate}
\end{equation}
so each of the 256 output dimensions draws from whichever branch carries the structure it needs. The gate is initialised with a bias towards the continuous branch to introduce another smoothness prior inspired by coarse-to-fine approaches. Coordinates pass through a fixed seam-free spherical lift of our own, which guarantees a non-degenerate longitude gradient at every orientation; see \citet{russwurm2024geographic} for the case against naive latitude--longitude parameterisations.

\paragraph{Time.} The time encoder maps the year, normalised over 1900--2035, to log-spaced Fourier features refined by a residual MLP~\citep{tancik2020fourier},
\begin{equation}
  \gamma(t) = \bigl[\sin(2\pi \bar{t}\, s f_k),\;
  \cos(2\pi \bar{t}\, s f_k)\bigr]_{k=1}^{32}, \qquad
  \boldsymbol{\tau} = \mathrm{N}_{32}\bigl(\mathrm{MLP}_{\mathrm{res}}
  (\gamma(t))\bigr),
  \label{eq:time}
\end{equation}
with a learned scale $s$: low frequencies carry decadal trend, high frequencies year-to-year structure, and the residual MLP supplies the nonlinear refinement the basis alone cannot. A complementary learned transport operator $T(\boldsymbol{\tau}, \Delta)$ represents signed temporal displacement; initialised near the identity, it is shaped by the consistency objectives of \S\ref{sec:objectives}.

\paragraph{Country and task.} A country identifier carries no geometric structure to exploit, so the country encoder is an embedding table refined by a residual MLP. The task encoder is an embedding table alone: the task vector conditions decoding jointly with the query's spatiotemporal state (\S\ref{sec:fusion}), and new rows allow unseen variables to be registered without disturbing trained tasks (\S\ref{sec:adaptation}). During training, \emph{embedding dropout} is applied to every encoder output before radius normalisation, regularising the embedding's direction while preserving the scale seen by fusion and alignment, and encouraging each latent variable to be useful for every training objective by preventing specialisation.

\subsection{Spatiotemporal state and conditional decoding}
\label{sec:fusion}
Geometry routing builds the shared spatial state $\mathbf{s} \in \mathbb{R}^{256}$: a coordinate query uses the location embedding directly, $\mathbf{s} = \mathbf{e}_\ell$, while a country query lifts the country embedding into the same space through a linear projector. Routing rather than mixing lets one fusion and decoding stack serve both geometries: after this point, every query is a 256-dimensional vector on a common bus, whatever its origin. Two cross-modal transformers~\citep{vaswani2017attention} then condition this state.

The first transformer injects time information into spatial embeddings. Its output is applied as a gated residual on the projected spatial state,
\begin{equation}
  \mathbf{z}_{st} = \mathrm{N}_{256}\!\bigl(\mathbf{s}' + g\,
  \boldsymbol{\delta}\bigr),
  \label{eq:stfusion}
\end{equation}
with the correction $\boldsymbol{\delta}$ zero-initialised and the gate $g$ starting small: fusion begins at the spatial identity and learns temporal structure only as it earns influence. The second transformer lets the spatiotemporal state query the task token. It queries $\mathbf{z}_{st}$, so its output (the conditioning
vector $\mathbf{c} \in \mathbb{R}^{512}$) is aware of when as well as where the query is observed.

A one-layer hypernetwork~\citep{ha2016hypernetworks} maps $\mathbf{c}$ to the weights of a small decoder, which is then applied to $\mathbf{z}_{st}$: the decoder is generated per query, its computation shaped jointly by task and spatiotemporal context. This is a more expressive form of task conditioning than concatenation or feature-wise modulation~\citep{rodriguez2025neural}, since each of the \numTasks{} tasks realises a different local decoding function, and makes adaptation to an unseen variable cheap.

\subsection{Predictive distribution and uncertainty}
\label{sec:uncertainty}
The generated decoder outputs the four parameters of a Normal--Inverse--Gamma distribution,
\begin{equation}
  \mu = a_\mu, \quad
  \nu = \mathrm{softplus}(a_\nu) + \tfrac{1}{2}, \quad
  \alpha = \mathrm{softplus}(a_\alpha) + 1 + 10^{-4}, \quad
  \beta = \mathrm{softplus}(a_\beta) + 10^{-4},
  \label{eq:nig}
\end{equation}
where the offsets guarantee the parameter constraints and the $\nu \ge \tfrac{1}{2}$ floor in particular prevents an epistemic blow-up as $\nu \to 0$. Because targets are standardised, a predictive width beyond a few standard deviations is a priori implausible; differentiable one-sided caps exploit this to bound both variance components without clamping gradients. Marginalising the latent Gaussian mean and variance yields a Student-$t$ predictive distribution with mean $\mu$, whose total variance admits the conventional moment decomposition
\begin{equation}
  \sigma^2_{\mathrm{pred}} =
  \underbrace{\frac{\beta}{\alpha - 1}}_{\sigma^2_{\mathrm{alea}}}
  \;+\;
  \underbrace{\frac{\beta}{\nu(\alpha - 1)}}_{\sigma^2_{\mathrm{epi}}}.
  \label{eq:decomposition}
\end{equation}
The two terms are useful read-outs: the first tracks observation and data variability, the second uncertainty in the predicted mean; but they are not perfectly separately identifiable from the marginal likelihood without additional assumptions~\citep{meinert2023unreasonable}. We therefore treat them as \emph{proxies}: we evaluate calibration of the total predictive distribution, and use the decomposition as a diagnostic rather than claiming separate calibration of its components. The total uncertainty also steers training, through an active task-resampling schedule.

\begin{figure}[t]
	\centering
	\includegraphics[width=0.9\linewidth]{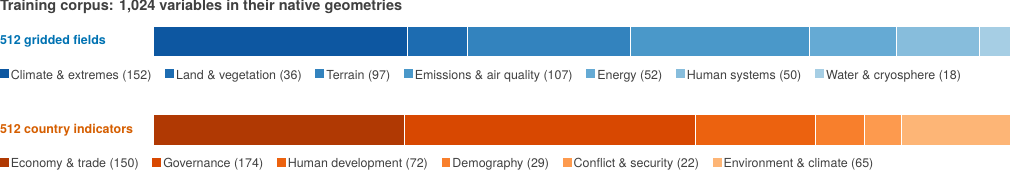}
	\caption{\textbf{Training corpus.} \numTasks{} variables in their native geometries:
		\numGridded{} gridded Earth-system fields and \numCountry{} country-level indicators.}
	\label{fig:corpus}
\end{figure}

\section{Training}\label{sec:training}

\subsection{Data and held-out evaluation}

TerraNova is trained on two reconstruction datasets and three fixed
alignment banks (Fig.~\ref{fig:corpus}). WorldTensor~\citep{rodriguez2026harmonised} contributes 512 fields on a common $0.25^{\circ}$ grid, covering, among others: climate, extremes, land use, vegetation, hydrology, energy, and human systems. A companion dataset we assemble for this work, which we name \textit{CountryTensor}, provides 512 national indicators selected from the Quality-of-Government ecosystem~\citep{teorell2021qog} for coverage and relevance, indexed by ISO3 code and year across development, institutions, governance, inequality, demography, and conflict. Both are historical observations rather than simulated data. The three alignment banks hold frozen image-derived embeddings (SatCLIP~\citep{klemmer2025satclip}, GeoCLIP~\citep{cepeda2024geoclip} and Copernicus-Embed~\citep{wang2025copernicus}) at 500{,}000 land coordinates generated by uniform Fibonacci sphere sampling~\cite{swinbank2006fibonacci}.
Targets are standardised per task, so one reconstruction objective serves heterogeneous units and scales and, as \S\ref{sec:uncertainty} exploits, unit marginal variance gives the evidential width caps a universal, task-comparable meaning. Further detail is given in \SIplace.

\subsection{Loss functions}
\label{sec:objectives}
Training interleaves one reconstruction family and three representation-level auxiliaries,
\begin{equation}
  \mathcal{L} = \lambda_{\mathrm{rec}} \mathcal{L}_{\mathrm{rec}} + \lambda_{\mathrm{tr}} \mathcal{L}_{\mathrm{tr}}
  + \lambda_{\mathrm{geo}} \mathcal{L}_{\mathrm{geo}} + \lambda_{\mathrm{cl}} \mathcal{L}_{\mathrm{cl}},
  \label{eq:total}
\end{equation}
with fixed weights $\lambda$ and a stochastically mixed step schedule (\S\ref{sec:optimisation}). The reconstruction term $\mathcal{L}_{\mathrm{rec}}$ is the exact negative log-likelihood of the NIG marginal plus two corrections that keep the evidential split well-posed on standardised targets: an evidence regulariser, which penalises standardised rather than raw error and helps avoid the evidence collapse of the original formulation~\citep{amini2020evidential, meinert2023unreasonable}, and the soft predictive-width cap of \S\ref{sec:uncertainty}.

\emph{Temporal transport} ($\mathcal{L}_{\mathrm{tr}}$) supervises the operator $T(\boldsymbol{\tau}, \Delta)$ simultaneously as a predictor and as a regulariser: a transported prediction must be accurate, transport must land on the true target-year embedding, zero displacement must be the identity, and semigroup consistency must be preserved. The consistency targets are stop-gradient, so these terms shape $T$ without collapsing the time encoder itself; the objective regularises temporal geometry rather than replacing direct prediction.

The two alignment objectives create TerraNova's coupled views of place, and both use sampled-negative InfoNCE~\citep{oord2018infonce}: for a positive pair $(\hat{\mathbf{s}}, \mathbf{v}^{+})$ and negatives
$\{\mathbf{v}^{-}_{k}\}$,
\begin{equation}
  \mathcal{L}_{\mathrm{align}} = -\log
  \frac{\exp\bigl(s(\hat{\mathbf{s}}, \mathbf{v}^{+})/\tau_{\mathrm{T}}\bigr)}
  {\exp\bigl(s(\hat{\mathbf{s}}, \mathbf{v}^{+})/\tau_{\mathrm{T}}\bigr)
  + \sum_{k} \exp\bigl(s(\hat{\mathbf{s}}, \mathbf{v}^{-}_{k})/\tau_{\mathrm{T}}\bigr)},
  \label{eq:infonce}
\end{equation}
with $s$ cosine similarity.

\emph{External geospatial alignment} ($\mathcal{L}_{\mathrm{geo}}$) projects the spatiotemporal state through a per-bank head into each embedding space and contrasts it with the bank vector at the same coordinate. The contrast is anchored in time: each bank is queried at the years its own training imagery was collected, so TerraNova's time-dependent state is always read in the bank's native temporal context; and biased towards hard negatives, drawn inverse-distance-weighted from nearby but distinct coordinates, which forces the representation to resolve fine spatial distinctions rather than global gradients alone. This objective distils image-derived geography without making imagery a target or inference-time input.

\emph{Country--location alignment} ($\mathcal{L}_{\mathrm{cl}}$) acts within TerraNova's own shared space: it aligns the country's spatiotemporal embedding with the mean embedding of coordinates sampled inside its territory, against other countries as negatives. Sampling density follows $\log(1 + P_y(\mathbf{x}))$ under the population for that year, so the territorial summary concentrates where people live (compressing the heavy-tailed population distribution so that dense cities do not monopolise it). This objective couples the two geometries directly; the external alignment provides the complementary image-derived anchor.

\subsection{Optimisation and active task resampling}
\label{sec:optimisation}
Each epoch interleaves fixed counts of reconstruction, transport and alignment steps in a globally shuffled schedule. Reconstruction sampling starts from the gridded/country mixture with per-domain floors and bounded task boosts, so the much larger gridded inventory does not crowd out the country indicators. The model trains for 128 epochs with AdamW~\citep{loshchilov2017adamw} under a warmup--cosine schedule~\citep{loshchilov2017sgdr}, gradient clipping, bf16-mixed precision, a batch size of 8{,}192, and an exponential moving average for evaluation. Every two epochs, \emph{active resampling} re-evaluates every reconstruction task on held-out points and updates its sampling probability from a smoothed mean of the total predictive variance (Eq.~\ref{eq:decomposition}): tasks the model is most uncertain about receive more training in the next interval. The multiplier is bounded so difficult tasks are boosted without starving the remainder.

\subsection{Adaptation to unseen tasks}
\label{sec:adaptation}
The task encoder reserves rows for variables absent from pretraining. For a new target $\tau^{\star}$, the native adaptation operator jointly optimises a fresh task row and rank-4 MiSS residuals~\citep{kang2024miss} in the task--spatiotemporal fusion: the input factors are layer-specific while a single output factor is shared across layers and sliced to each map's width, for roughly $4.2 \times 10^{5}$ trainable parameters in total. All pretrained weights and trained task rows remain frozen.

\begin{figure}[t]
\centering
\begin{subfigure}[b]{0.375\linewidth}
\centering
\includegraphics[width=\linewidth]{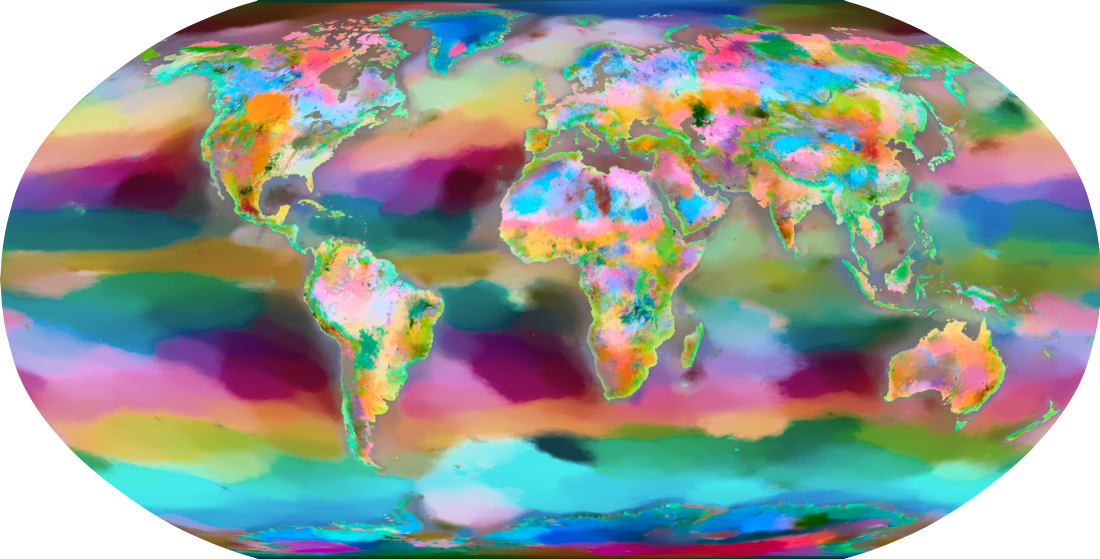}
\caption{Location embedding.}
\label{fig:anatomy-robinson}
\end{subfigure}
\hfill
\begin{subfigure}[b]{0.272\linewidth}
\centering
\includegraphics[width=\linewidth]{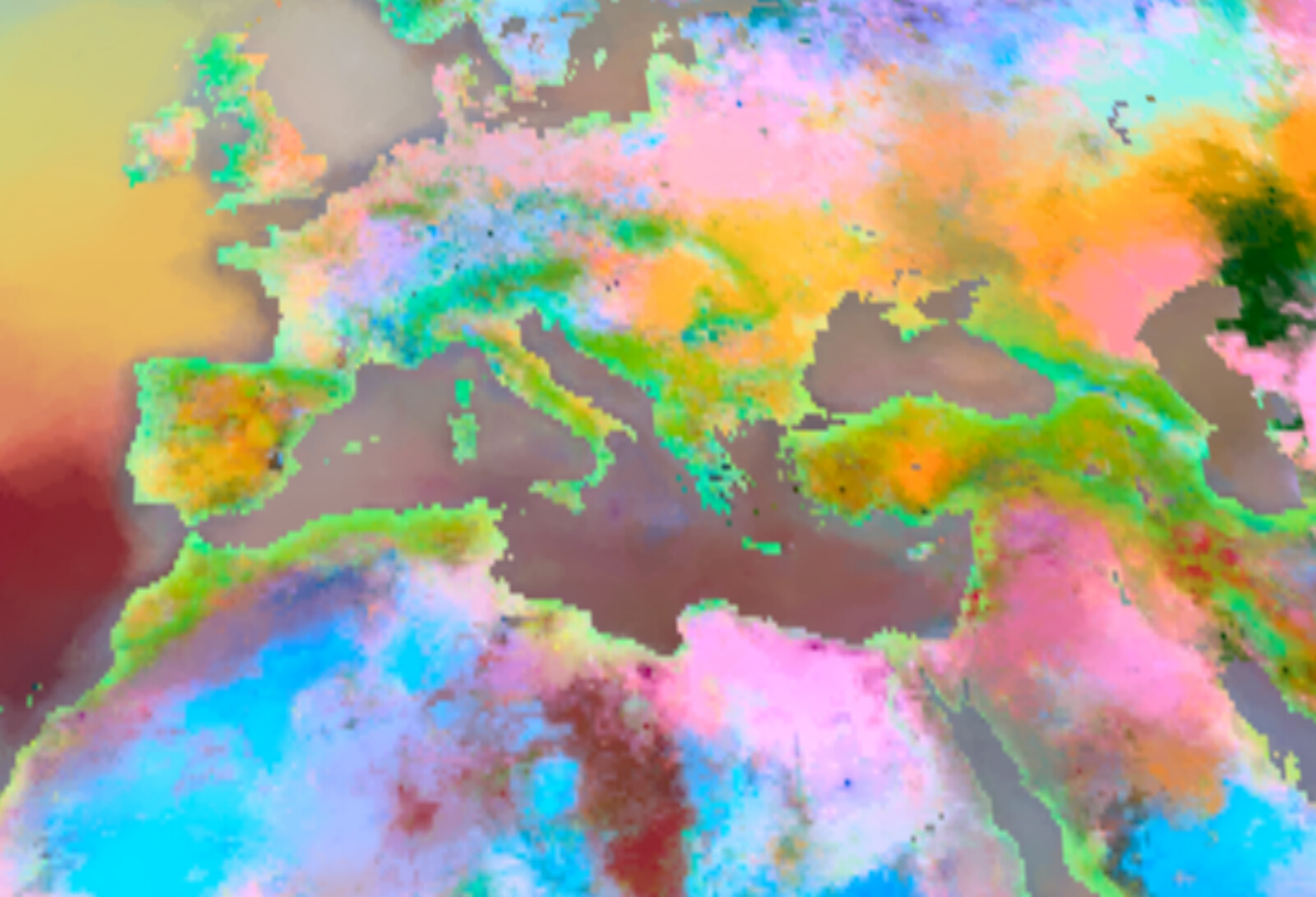}
\caption{A local zoom.}
\label{fig:anatomy-inset}
\end{subfigure}
\hfill
\begin{subfigure}[b]{0.272\linewidth}
\centering
\includegraphics[width=\linewidth]{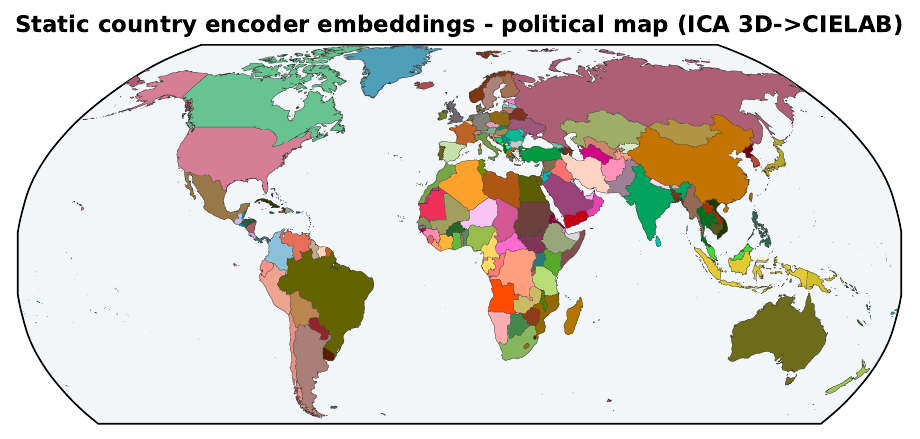}
\caption{The country embeddings.}
\label{fig:anatomy-political}
\end{subfigure}
\caption{\textbf{(a)}~The location embedding on a Robinson projection (PCA, then UMAP~\cite{mcinnes2018umap} to three dimensions, mapped into CIELAB~\cite{luo2023cielab}). \textbf{(b)}~The same at a local zoom. \textbf{(c)}~The static country embeddings (ICA to three dimensions, then CIELAB).}
\label{fig:anatomy}
\end{figure}
\vspace{-2mm}

\begin{figure}[t]
\centering
\begin{subfigure}[b]{0.42\linewidth}
\centering
\includegraphics[width=\linewidth]{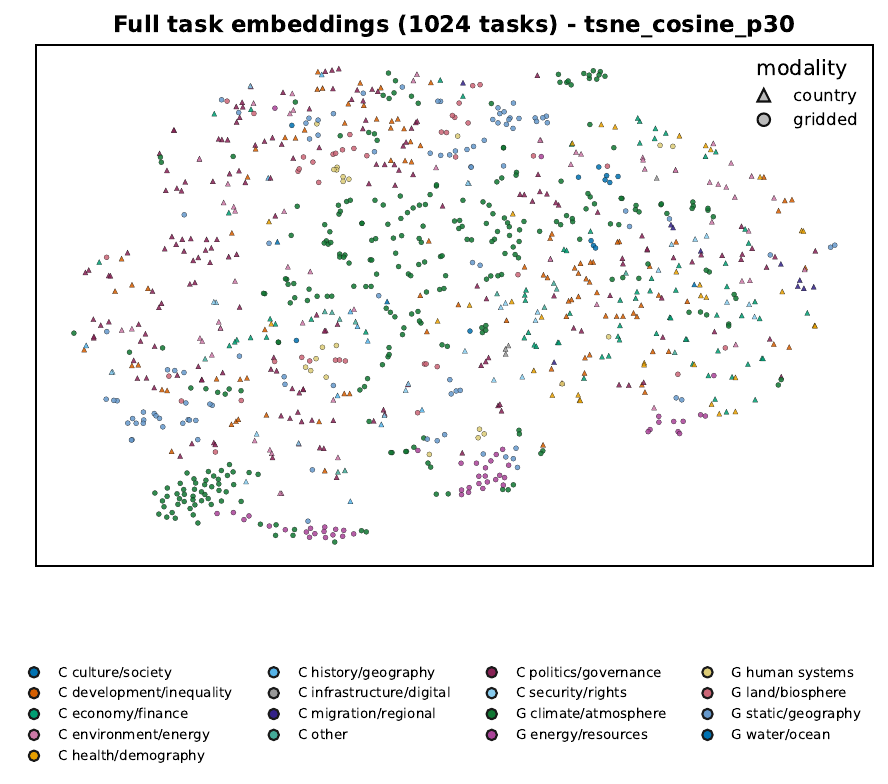}
\caption{Task embeddings (t-SNE).}
\label{fig:anatomy-tasks}
\end{subfigure}
\hfill
\begin{subfigure}[b]{0.47\linewidth}
\centering
\includegraphics[width=\linewidth]{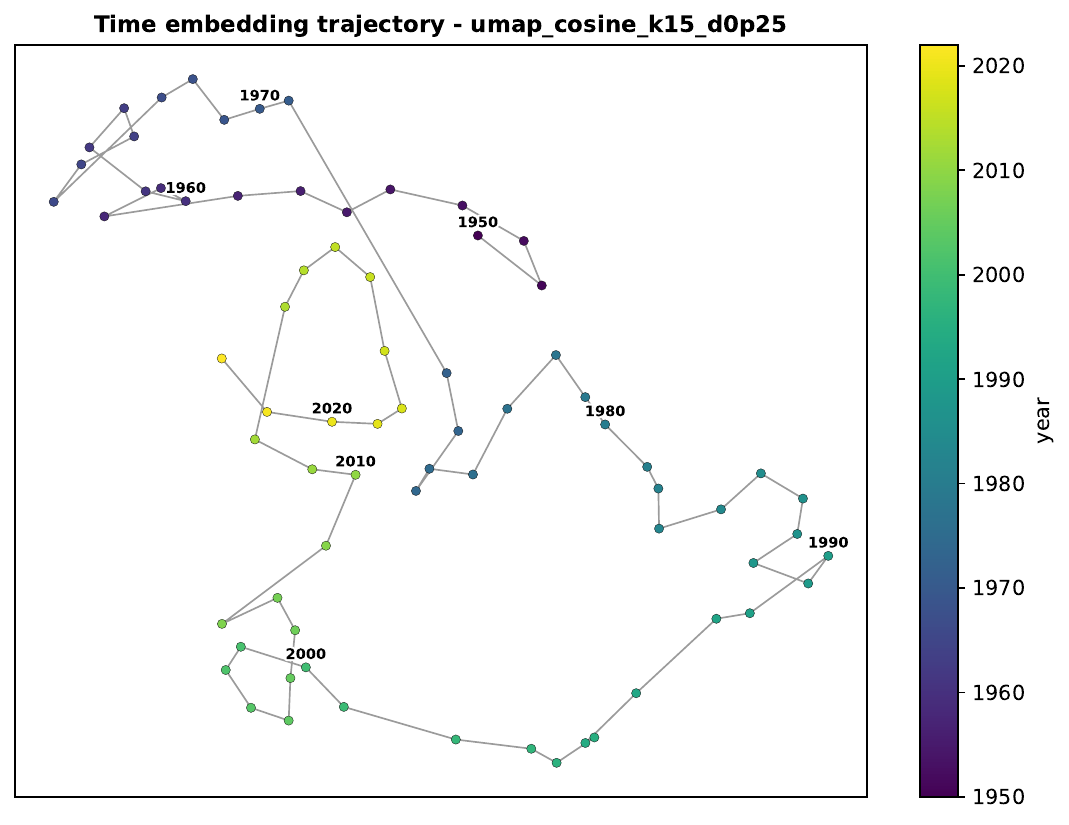}
\caption{Time embeddings (UMAP).}
\label{fig:anatomy-time}
\end{subfigure}
\caption{\textbf{(a)}~A t-SNE~\citep{maaten2008visualizing} of the \numTasks{} task embeddings. \textbf{(b)}~A UMAP of the year embeddings, 1950 to 2020.}
\label{fig:anatomy_spaces}
\end{figure}

\section{Validation}\label{sec:validation}

In this section, we evaluate the model design and the learned representations. We test three properties: that the latent space is organised rather than memorised, that the coupling between the two geometries is real and separable into distinct contributions, and that the per-query uncertainty is trustworthy. Please refer to \SIplace{} for extensive per-component ablations.

\subsection{Embedding analysis}\label{sec:anatomy}

The encoders recover structure that was never explicitly provided as a target. Projecting the location encoder's output over the globe (Fig.~\ref{fig:anatomy}) recovers land topography, coastal areas, biome boundaries and coherent ocean basins, and a local zoom shows the same organisation at fine detail. The static country encoder, rendered as a political choropleth with each country coloured by a dimensionality reduction of its embedding, sorts by region. In their own spaces (Fig.~\ref{fig:anatomy_spaces}) the task embeddings cluster by scientific domain while separating the gridded fields from the country indicators, and the time encoder's year embeddings trace a smooth, ordered trajectory from 1950 to 2020. The organisation is continuous: nearby places, regions, domains and years receive nearby codes, and the principal axes of variation align with real geography. The manifold has an intrinsic dimension~\citep{rao2025measuring} of \idTerranova{} (FisherS estimator, across land and land--ocean pools).

\paragraph*{The representation moves through time}\label{sec:repr-time}

\begin{figure}[t]
\centering
\includegraphics[width=.5\linewidth]{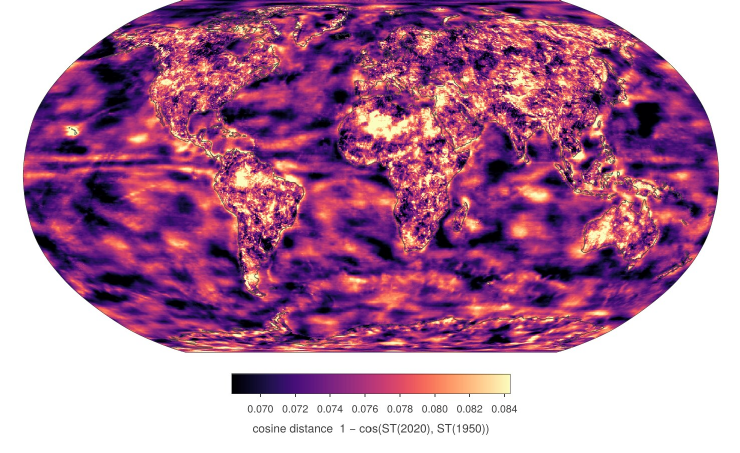}
    \caption{Cosine distance between each place's spatiotemporal representation in 2020 and in 1950.}
\label{fig:repr-time}
\end{figure}

Because the state is spatiotemporal, the representation of a fixed place varies across years. The per-cell cosine distance $1-\cos(\mathrm{ST}(2020),\mathrm{ST}(1950))$ between each place's representation in 2020 and in 1950 (Fig.~\ref{fig:repr-time}) is largest over populated and developing areas and locations with rapidly changing climate; and smallest over the stable interiors of the oceans, so the change in the representation is concentrated where the observed world changed most, with no explicit change target.

\paragraph*{An emergent development axis}\label{sec:development}

\begin{figure}[t]
\centering
\begin{subfigure}[b]{0.45\linewidth}
\centering
\includegraphics[width=\linewidth]{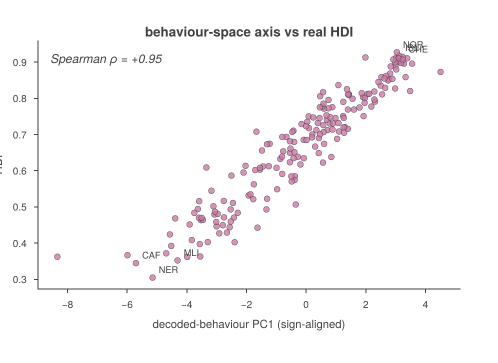}
\caption{Decoded behaviour.}
\label{fig:development-decoded}
\end{subfigure}
\hfill
\begin{subfigure}[b]{0.47\linewidth}
\centering
\includegraphics[width=\linewidth]{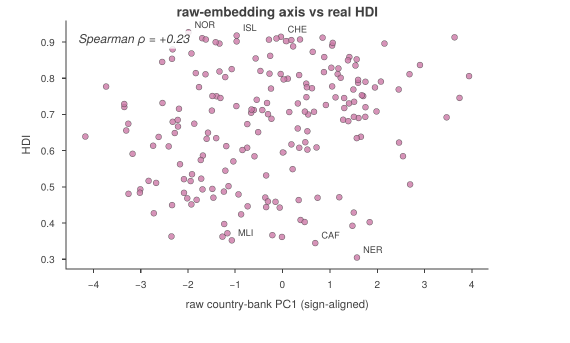}
\caption{Raw embedding.}
\label{fig:development-raw}
\end{subfigure}
\caption{\textbf{An emergent development axis.} The leading principal component of the country representation against the real Human Development Index, which enters neither decomposition. \textbf{(a)}~From the \emph{decoded} indicator panel. \textbf{(b)}~From the \emph{raw} embedding vectors.}
\label{fig:development}
\end{figure}

The embeddings also carry socioeconomic structure, and the construction that recovers it runs through the decoder rather than over the embeddings themselves. We take the learned embedding of each of the 246 countries the model was trained on, decode it through the model's own task-conditioned head into the values it predicts for eleven curated national indicators in 2015, and assemble one row per country. Each column is then standardised across countries, so that indicators on different units contribute comparably, and the Human Development Index and GDP per capita are dropped from the panel before any decomposition, leaving \devPanel{} indicators: life expectancy, infant and under-five mortality, fertility, urbanisation, internet adoption, the Gini index, carbon footprint and liberal democracy. The leading principal component of that $246\times9$ matrix carries \devEvr{} of its variance and tracks the real Human Development Index at Spearman $|\rho|\approx\devHDI{}$ and real log GDP per capita at $\approx\devGDPpc{}$ (Fig.~\ref{fig:development}a), neither of which entered the decomposition. The sign of a principal component is arbitrary, and the figure orients the axis so that it increases with development. Three indicators that remain in the panel, life expectancy and the two mortality rates, bear on the longevity dimension HDI is itself built from, so we repeat the decomposition with them removed. Dropping life expectancy leaves the correlation at $\devHDInoLife{}$, and dropping the two mortality rates as well, which reduces the panel to the six remaining indicators of demography, infrastructure, inequality, environment and institutions, leaves it at $\devHDInoLongevity{}$. The panel contains no education variable at all, so HDI's third dimension is absent throughout. The same construction applied to the raw 64-dimensional country vectors gives a flat spectrum, with a leading component carrying \devRawEvr{} of the variance and correlating with HDI at only $\devRawHDI{}$ (Fig.~\ref{fig:development}b). The development gradient is therefore carried by what the model \emph{predicts} about each country, recovered from the joint structure it has learned across hundreds of national indicators, rather than by the leading direction of variation in the raw embedding.

\paragraph*{Embedding arithmetic}\label{sec:arithmetic}

\begin{figure}[t]
\centering
\includegraphics[width=.7\linewidth]{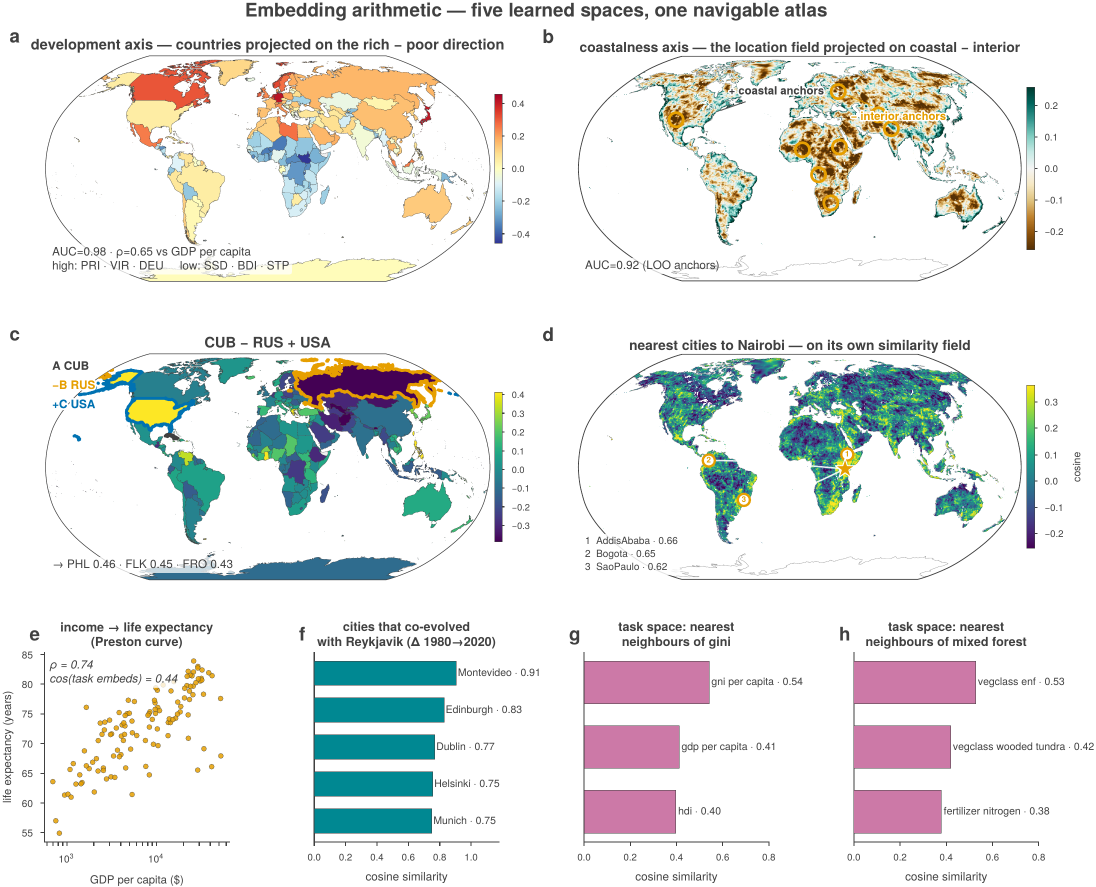}
\caption{Embedding arithmetic across the model spaces: linear semantic axes, a composed country analogy, nearest-neighbour queries in each space, and a decoded relationship read across spaces.}
\label{fig:arithmetic}
\end{figure}

The learned spaces support vector operations analogous to word embeddings~\citep{mikolov2013linguistic,almeida2019word} (Fig.~\ref{fig:arithmetic}). Linear directions carry meaning: projecting the country field onto a rich-minus-poor direction recovers the development axis (AUC $0.98$, $\rho\approx0.65$ against GDP per capita), and projecting the location field onto a coastal-minus-interior direction recovers a coastalness axis (AUC $0.92$). Analogies compose in the country space: Cuba minus Russia plus the United States lands near the Philippines, the Falkland Islands and the Faroe Islands. Nearest-neighbour queries return sensible neighbours in all three spaces: the coordinates most similar to Nairobi are Addis Ababa, Bogot\'a and S\~ao Paulo; the cities whose representations co-evolved most with Reykjav\'{\i}k between 1980 and 2020 are Montevideo, Edinburgh and Dublin; and in task space the nearest neighbours of the Gini index are GNI per capita, GDP per capita and the Human Development Index, while those of mixed-forest cover are other vegetation classes. Across spaces, decoded income and life expectancy trace the empirical Preston curve ($\rho\approx0.74$). These examples indicate that structure in these spaces is accessible through directions and offsets constructed by hand, which supports use of the representation for exploration as well as prediction.

\subsection{Coupling between geometries}\label{sec:coupling}

We next test whether the coupling between the two geometries, the population-weighted alignment that ties each country to the field over its territory (\S\ref{sec:objectives}), is effective. We first test cross-geometry retrieval: we take the mean embedding of the coordinates inside a country and ask which of a held-out pool of \retrievalPool{} countries it names. TerraNova reaches recall@1 \retrievalRecallOne{}, rising to \retrievalRecallFive{} at recall@5 and \retrievalRecallTen{} at recall@10. Since the query is a territorial summary rather than a single coordinate, what this measures is whether a country's own code sits where its territory sits. It is a question the geospatial baselines cannot pose at all, since each holds only one geometry, and passing it is evidence that the alignment placed the two geometries in one common latent space rather than leaving them in two unrelated ones. The coupling is soft throughout, encouraged rather than imposed.

\begin{figure}[t]
\centering
\includegraphics[width=.8\linewidth]{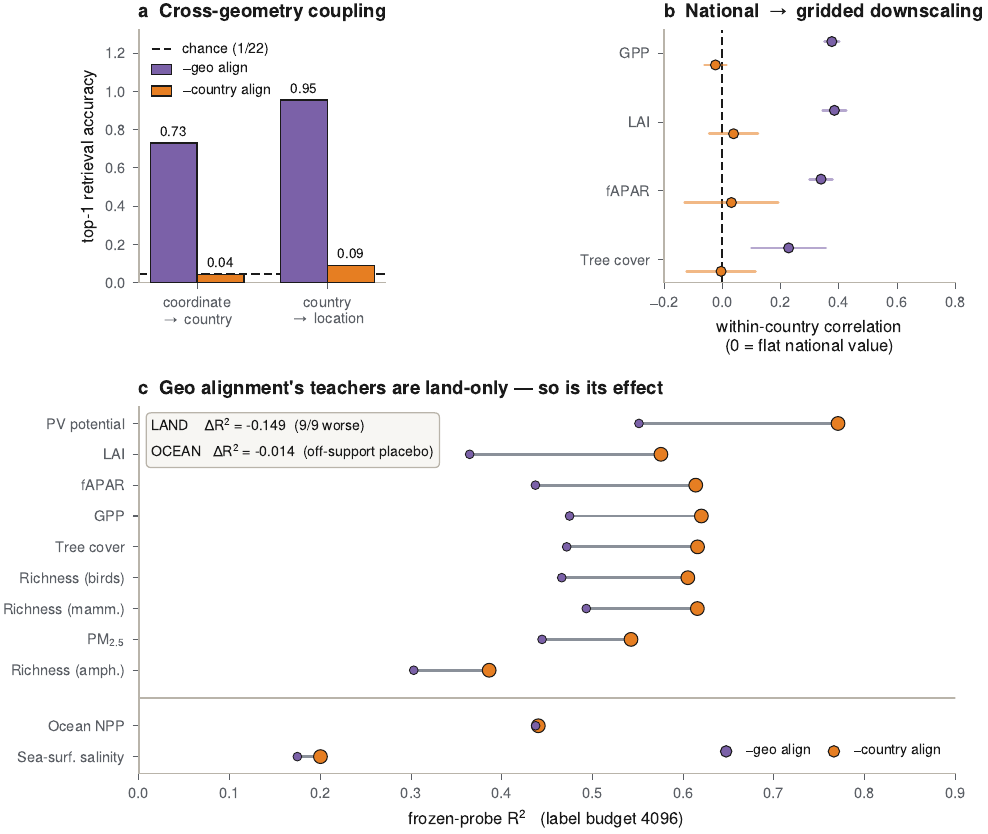}
\caption{A crossed pair of leave-one-out models, identical except for which alignment objective each drops. \textbf{(a)}~Cross-geometry retrieval in both directions, read from the native embeddings with no learned probe, over a pool of \retrievalPoolAbl{} countries. \textbf{(b)}~National-to-gridded downscaling on four biosphere targets. \textbf{(c)}~Frozen-probe reconstruction $R^2$, split into land and ocean targets.}
\label{fig:loo-alignment-main}
\end{figure}

To show that the two alignment objectives do separate work, we run an ablation study: a crossed pair of leave-one-out experiments, one dropping the country--location alignment from training, and one dropping the geospatial alignment, otherwise identical, so that each serves as the other's matched control (Fig.~\ref{fig:loo-alignment-main}). Removing the internal alignment removes the cross-geometry capabilities and only those: coordinate-to-country retrieval collapses to chance ($0.043$) and national-to-gridded downscaling falls to the flat national-mean null, while reconstruction and calibration are untouched (RMSE $0.276$ against $0.272$, expected calibration error $0.173$ against $0.172$); the two geometries are decoupled, not degraded. Removing the external alignment instead costs representation quality, and it does so on the support of the teacher banks it borrows from: $-0.149$ $R^2$ on all nine \emph{land} targets and essentially nothing ($-0.014$) on the two \emph{ocean} targets. Because the three teacher banks are land-only by construction, the ocean targets act as a support-matched placebo. One objective builds the common space; the other enriches it. This double dissociation is a methodological finding of the paper: two objectives drawing on different observations of one world contribute different, separable structure to the representation.

\subsection{Uncertainties}\label{sec:atlas}

\begin{figure}[t]
\centering
\includegraphics[width=.68\linewidth,trim=0 18 0 0,clip]{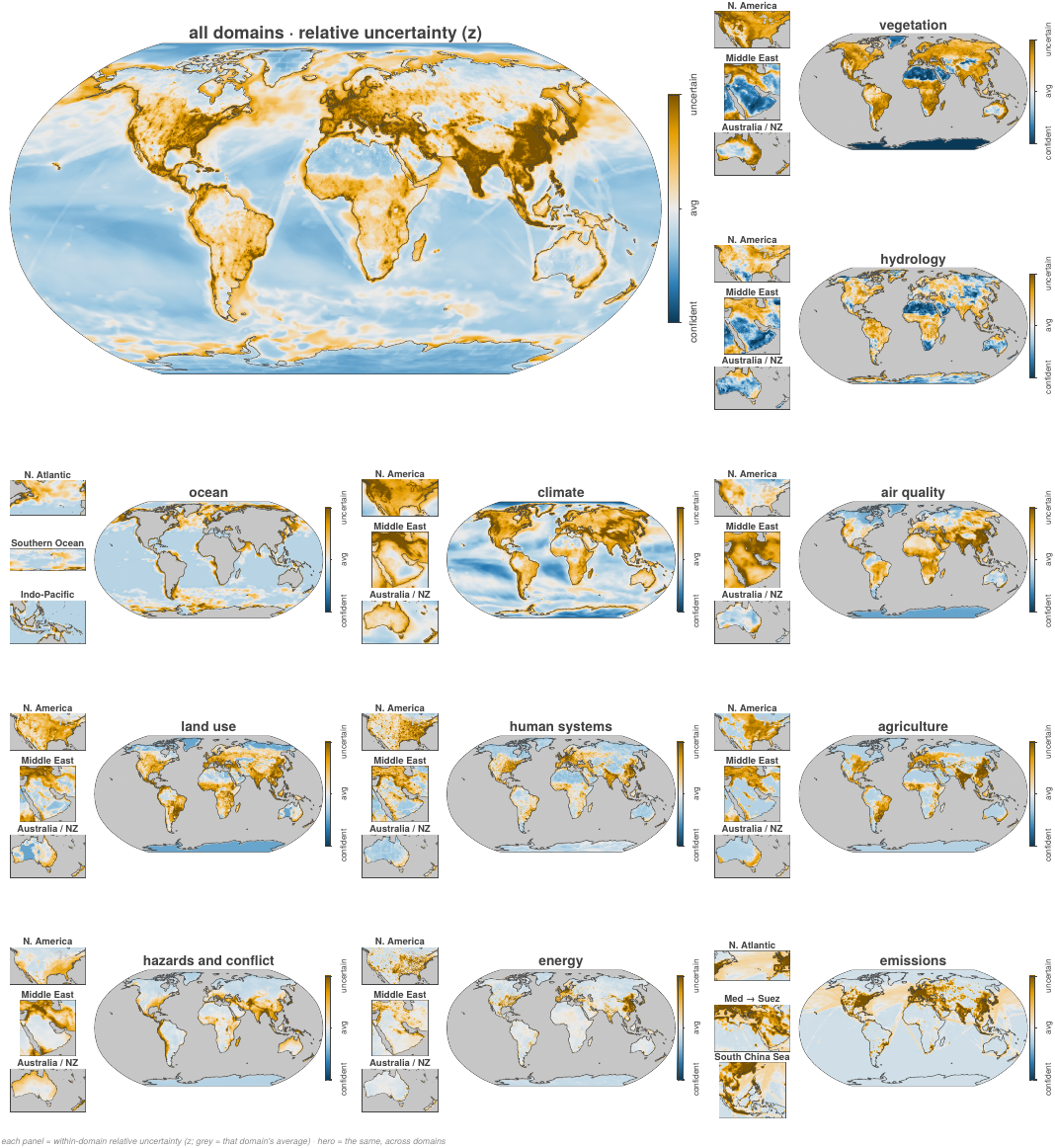}
\caption{Per-domain maps of predictive uncertainty, each standardised within its own domain.}
\label{fig:atlas}
\end{figure}

Every TerraNova query returns a NIG predictive distribution rather than a point estimate, and how far that distribution can be trusted depends on which of two properties is asked for. Its ordering is informative and its scale is not. On unseen variables the raw evidential interval is systematically too wide, which we correct with one recalibration constant per task: a country-split conformal interval restores nominal $90\%$ coverage on held-out years. Every interval we report is recalibrated. Mapping the predictive uncertainty per domain (Fig.~\ref{fig:atlas}), standardised within each domain so that domains on very different scales can be compared, gives an atlas of where the model is least certain. Its geography is one of difficulty rather than of coverage: the width is largest where a field is strongly structured, over China, western Europe and the eastern United States for emissions, over the Congo and Amazon basins for vegetation and over the Eurasian interior for climate, and it is smallest over more homogeneous places such as the Sahara, the Australian desert and the open subtropical gyres. 

\section{Results}\label{sec:results}

In this section, we measure what TerraNova delivers on held-out data: static prediction against purpose-built geospatial encoders (\S\ref{sec:benchmark}), reconstruction and nowcasting of national records through time (\S\ref{sec:nowcast}), dense fields from sparse measurements (\S\ref{sec:sparse}), national cross-sections from a few reporting countries (\S\ref{sec:recon-country}), national-to-gridded downscaling (\S\ref{sec:downscale}) and what all of this costs to train and to run (\S\ref{sec:cost}). Splits are fixed across methods and label budgets, regression is scored by held-out $R^2$ and Pearson correlation, classification by accuracy, and predictive distributions by coverage and sharpness.

\subsection{Comparison with existing geospatial embeddings}\label{sec:benchmark}

\begin{figure}[t]
\centering
\includegraphics[width=0.9\linewidth]{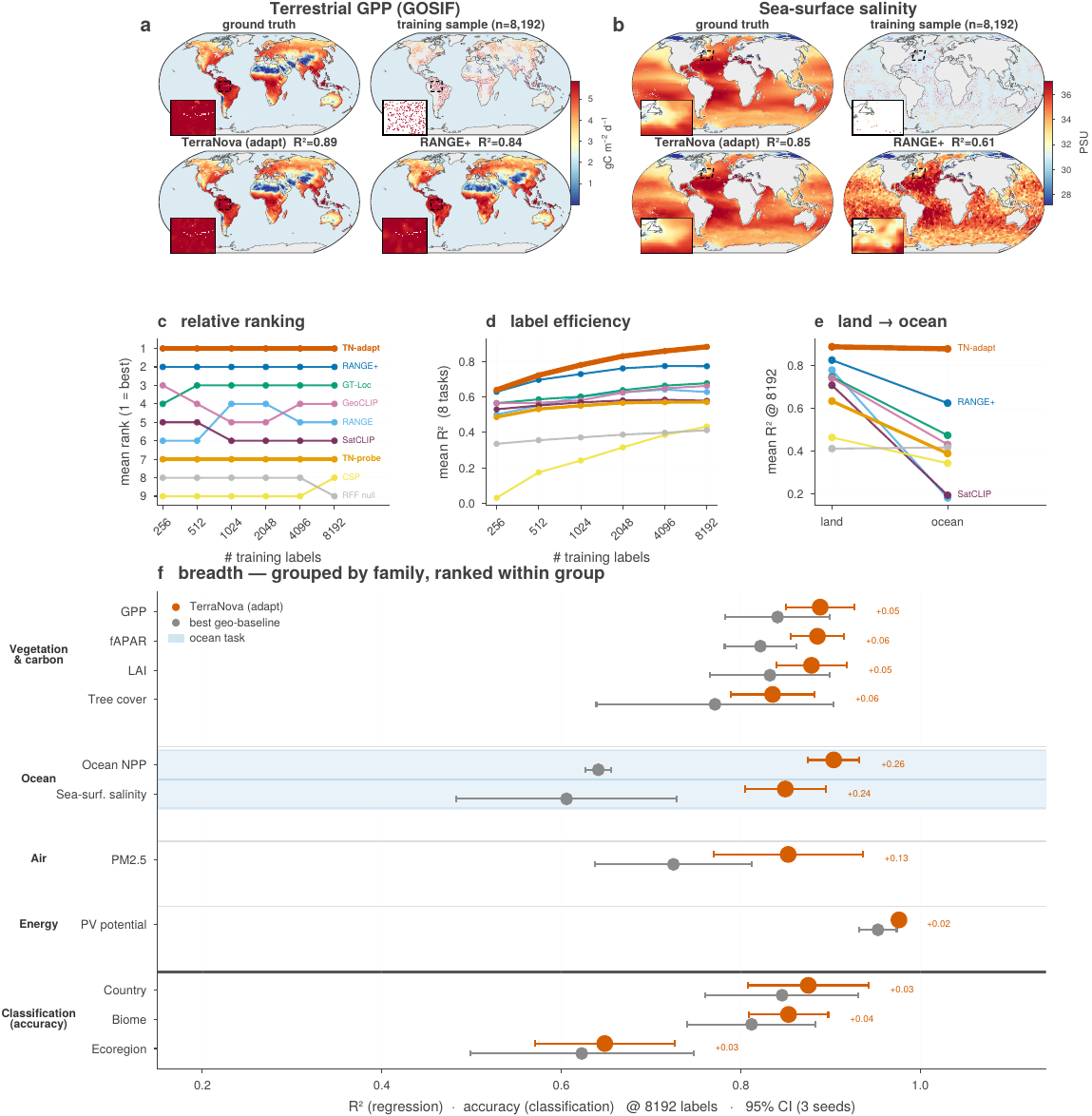}
\caption{\textbf{Comparison with existing geospatial embeddings, on unseen data.} \textbf{a},\textbf{b}, Two targets at $8{,}192$ labels: truth, training sample, TerraNova's reconstruction and the strongest baseline. \textbf{c}, Mean rank against label budget, where TN-adapt is the decoder read-out and TN-probe the frozen-embedding read-out of the same model. \textbf{d}, Mean held-out $R^2$, and \textbf{e}, the same split into land and ocean. \textbf{f}, Per-target results against the best baseline for that target.}
\label{fig:benchmark}
\end{figure}

In this benchmark, we hold out eleven environmental and eco-geographic targets, eight scored by $R^2$ and three by accuracy, and compare nine read-outs at label budgets from $256$ to $8{,}192$ over three seeds (Fig.~\ref{fig:benchmark}). Six models are published location encoders read out by a probe on their frozen embeddings, one is a random-Fourier-feature null, and two are TerraNova: a probe on its frozen embedding, which is the like-for-like comparison, and its own decoder with a fresh task row and a rank-$4$ MiSS adapter, the recommended reuse path. The adapted read-out receives coordinates only and no time, so its margin cannot come from temporal information the baselines cannot express.

MiSS-adapted TerraNova reaches a mean held-out $R^2$ of \benchMeanR{} against \benchMeanRBase{} for the strongest baseline (RANGE+~\cite{dhakal2025range}), and it is at or above the best geospatial baseline on every target, by \benchDeltaRange{} (Fig.~\ref{fig:benchmark}d,f). The ordering depends on the label budget only at the low end, where the adapter has little to fit: by mean rank RANGE+ leads at $256$ labels, and from \benchCrossover{} upwards TerraNova is best at every budget (Fig.~\ref{fig:benchmark}c). What delivers that lead is the cheaply-adaptable decoder rather than the raw geometry. Probed linearly, TerraNova's own embedding ranks seventh of the nine, and the margin appears only once a fresh task row and rank-$4$ MiSS residuals are fitted, at \adaptParams{} trainable parameters per task. The margin is widest where the baselines have no training signal. On the two marine targets TerraNova reaches \benchOceanTN{}, while the image-trained encoders lose \benchOceanDrop{} $R^2$ between land and ocean and the best of them reaches \benchOceanBase{} (Fig.~\ref{fig:benchmark}e). Time is another axis these encoders cannot express, and we explore it next.

\subsection{Temporal interpolation and nowcasting}\label{sec:nowcast}

\begin{figure}[htbp]
\centering
\includegraphics[width=0.9\linewidth]{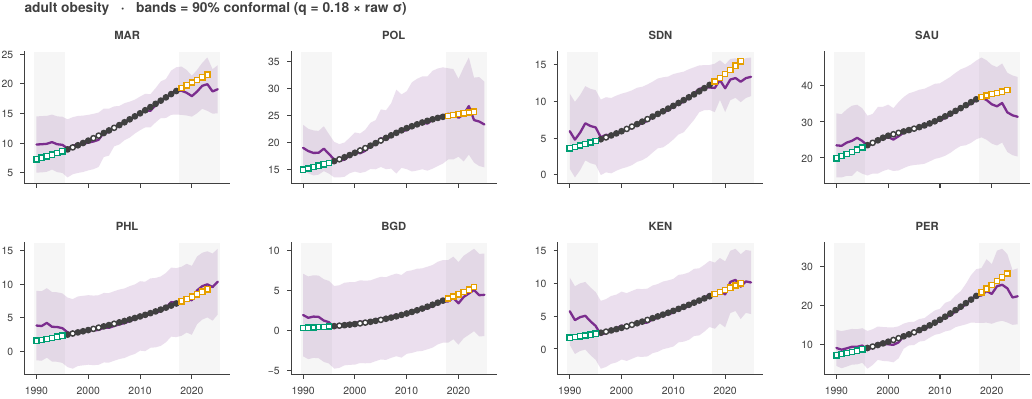}
\caption{\textbf{Reconstructing and nowcasting.} Filled circles are training years, open circles held-out years inside the window and open squares the held-out blocks at either end; the line is the predicted mean and the band the conformal $90\%$ interval.}
\label{fig:temporal}
\end{figure}

\begin{figure}[t]
\centering
\includegraphics[width=0.8\linewidth]{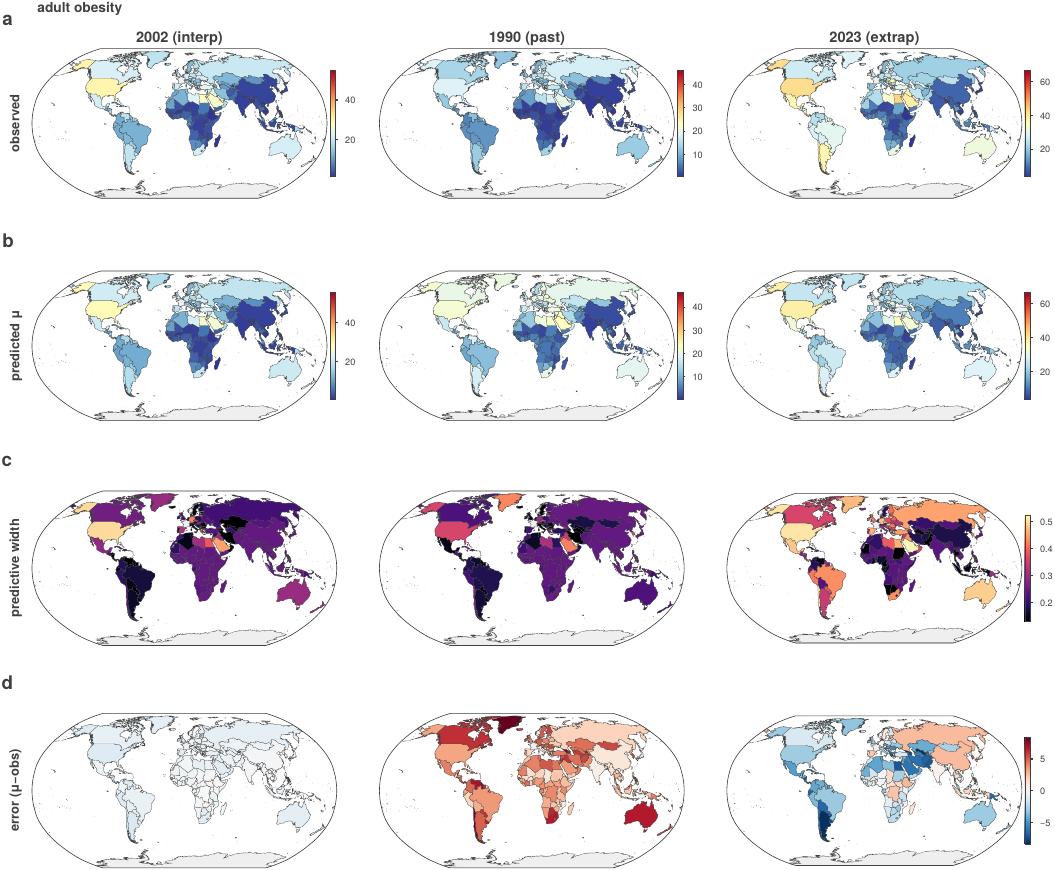}
\caption{The fit of Fig.~\ref{fig:temporal} read out over every country at three held-out years. Rows are \textbf{a}, the observation, \textbf{b}, the predicted mean, \textbf{c}, the predictive width and \textbf{d}, the error; columns are an interpolated year, a year before the record starts and a year after it ends.}
\label{fig:temporal_maps}
\end{figure}

In this experiment, we aim to assess the temporal capabilities of TerraNova. Each national record (unseen during training) is split by holding out its earliest and latest years, and the model is adapted on the years in between by fitting a fresh task row together with rank-$4$ MiSS residuals on the frozen backbone, then asked to predict both held-out ends (Fig.~\ref{fig:temporal}). Across the battery of \numTemporalTasks{} indicators the mean short-range anomaly correlation is \medianAnom{} and the median \shortRangeMedian{}, ranging from near-perfect on smooth series to near-zero on intrinsically noisy ones. Skill is highest at the training-window edge (\anomEdge{}) and decays with lead time, backcasts holding \anomBackcastFar{} six years into the past and forecasts \anomForecastFar{} six years ahead. A time-aware trend extrapolation is a closer reference: TerraNova leads it at the training-window edge, matches it throughout the backcast direction, and falls behind it in the forecast direction beyond about two years. The evidential interval over-covers on the held-out years, which we correct with country-split conformal recalibration. The same fit reconstructs three national cross-sections from one adapter (Fig.~\ref{fig:temporal_maps}). The between-country pattern is recovered almost exactly in an interpolated year (level correlation \mapsLevelInterp{}, mean absolute error \mapsErrInterp{} percentage points) and holds at both extrapolated ends (\mapsLevelExtrap{}, \mapsErrExtrap{} points), and the predictive width is \mapsWidthGrow{} wider at the forecast year than at the interpolated one. One frozen model serves indicators as different as scientific output, tuberculosis incidence, adult obesity and patent applications.

\subsection{Gridded reconstruction from sparse measurements}\label{sec:sparse}

\begin{figure}[t]
\centering
\includegraphics[width=0.9\linewidth]{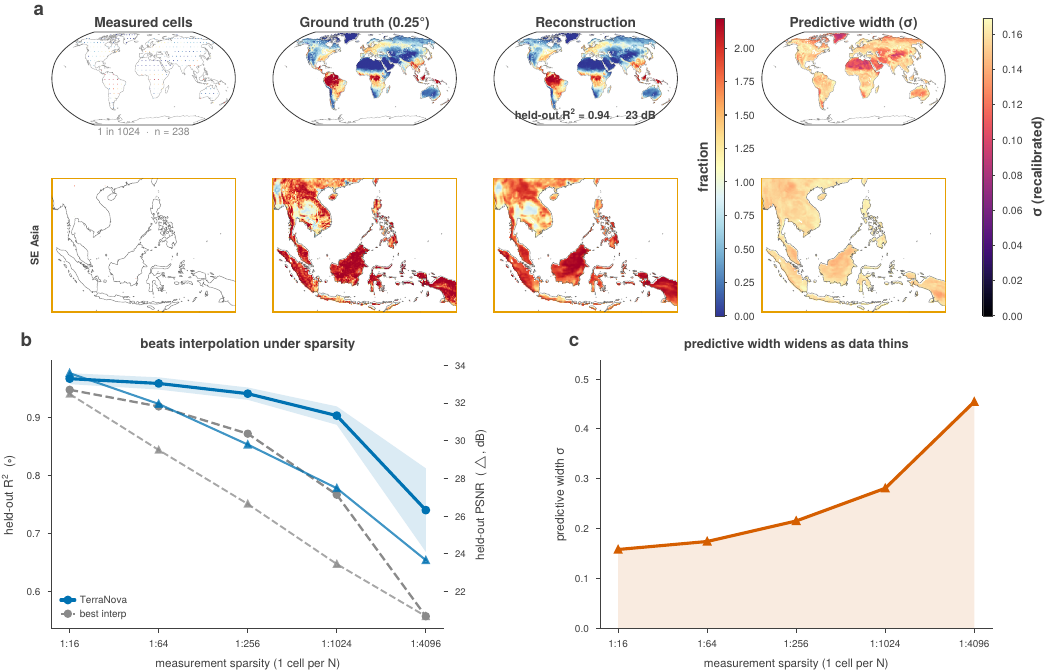}
\caption{\textbf{a}, A dense $0.25^\circ$ field reconstructed from a sparse lattice of measured cells (measured $\mid$ truth $\mid$ reconstruction $\mid$ predictive width~$\sigma$). \textbf{b}, Held-out $R^2$ (circles, left axis) and PSNR (triangles, right axis) against measurement sparsity, for the best adaptation read-out and the best classical interpolator. \textbf{c}, The predictive width as a fraction of the target's standard deviation.}
\label{fig:sparse}
\end{figure}

We now assess whether TerraNova can reconstruct a dense $0.25^\circ$ field from a sparse lattice of measured cells, for variables it never saw in training. Parameter-efficient adaptation of the frozen backbone holds up at measurement densities where classical interpolation degrades sharply. From \sparseHeroFrac{} of the cells, roughly one measurement every eight degrees, the target shown is recovered at a held-out $R^2$ of \sparseHeroRtwo{} and \sparseHeroPSNR{}~dB, scored under the held-out new-region split of \S\ref{sec:training} (Fig.~\ref{fig:sparse}a). Pooled over the leakage-free targets, each scored with the best of six adaptation read-outs against the best of five interpolators, the two approaches are close where measurements are dense (\sparseRtwoDense{} against \sparseInterpDense{} at one cell in sixteen) and the margin widens monotonically as the lattice thins, reaching \sparseRtwoSparse{} against \sparseInterpSparse{} at one cell in $4{,}096$ (Fig.~\ref{fig:sparse}b). Unlike interpolation, the model attaches a per-cell predictive width to every value it fills in, widening from \sigmaWiden{} of the target's standard deviation as the lattice thins (Fig.~\ref{fig:sparse}c), and it stays calibrated as it widens.

\subsection{Country-level extrapolation}\label{sec:recon-country}
\begin{figure}[t]
\centering
\includegraphics[width=0.9\linewidth]{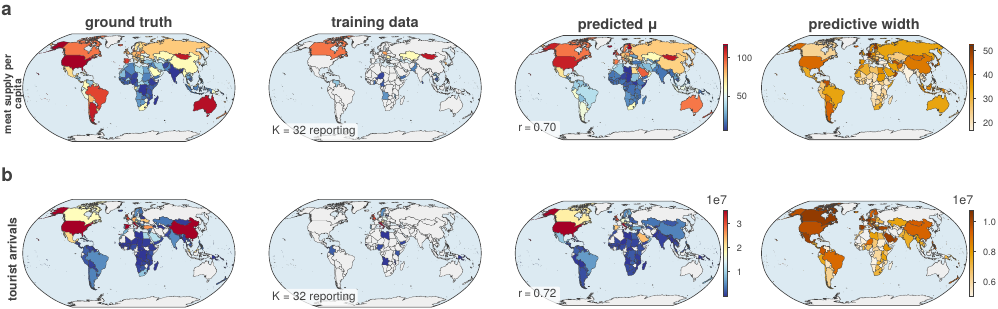}
\caption{\textbf{Few nations in, all nations out.} Two indicators reconstructed from \reconKdisp{} reporting countries: truth, the revealed countries, the predicted national level and the predictive width. \textbf{a}, Meat supply per capita. \textbf{b}, Tourist arrivals. Maps show one seed; the text quotes three-seed means.}
\label{fig:recon-country}
\vspace{-6mm}
\end{figure}
We use the same operator to reconstruct a national cross-section from a few reporting countries. For each held-out indicator we keep its best-covered year, hide $30\%$ of countries, reveal the value for $K$ of the rest, fit a fresh task row with rank-$4$ MiSS residuals \citep{kang2024miss} on the country route, and predict the hidden national levels. At $K=\reconKdisp{}$, roughly one country in seven, held-out level correlation reaches \reconMeat{} for meat supply per capita and \reconTour{} for tourist arrivals (Fig.~\ref{fig:recon-country}), and pooled over the \reconNumTasks{} indicators of the battery it climbs across \reconTnCurve{} as the reporting budget grows from four countries to $128$. A frozen-embedding probe of the same cross-sections trails the adapted read-out at every budget. The natural baseline is spatial interpolation between the reporting countries, and the strongest of the four tried is inverse-distance weighting over country centroids, at \reconIdwCurve{} over the same budget range: a handful of points spread over the globe is too little to fit a Gaussian-process length scale. The learned representation and spatial interpolation carry different information about a national cross-section. What separates them is spatial smoothness: TerraNova's margin over interpolation anti-correlates with how well interpolation itself does ($r=\reconDissoc{}$ across indicators), so the representation is far ahead where a country's value is not predictable from its neighbours, as for tourist arrivals (\reconTour{} against \reconTourBase{}), and behind where it is. Their held-out errors decorrelate as coverage grows, from \reconErrCorrHigh{} at four reporting countries to \reconErrCorrLow{} at $128$. Neither arm dominates: TerraNova leads at four, sixteen, sixty-four and $128$ revealed countries, interpolation leads at eight and thirty-two, and per indicator at $K=\reconKdisp{}$ TerraNova is ahead on \reconTnWins{} of \reconNumTasks{}. We therefore report the reconstruction as a capability of the coupled representation and the combination as the way to use it, not as a claim on spatial statistics' own ground.

\subsection{Gridded field estimation from country-level values}\label{sec:downscale}

\begin{figure}[t]
\centering
\includegraphics[width=0.9\linewidth]{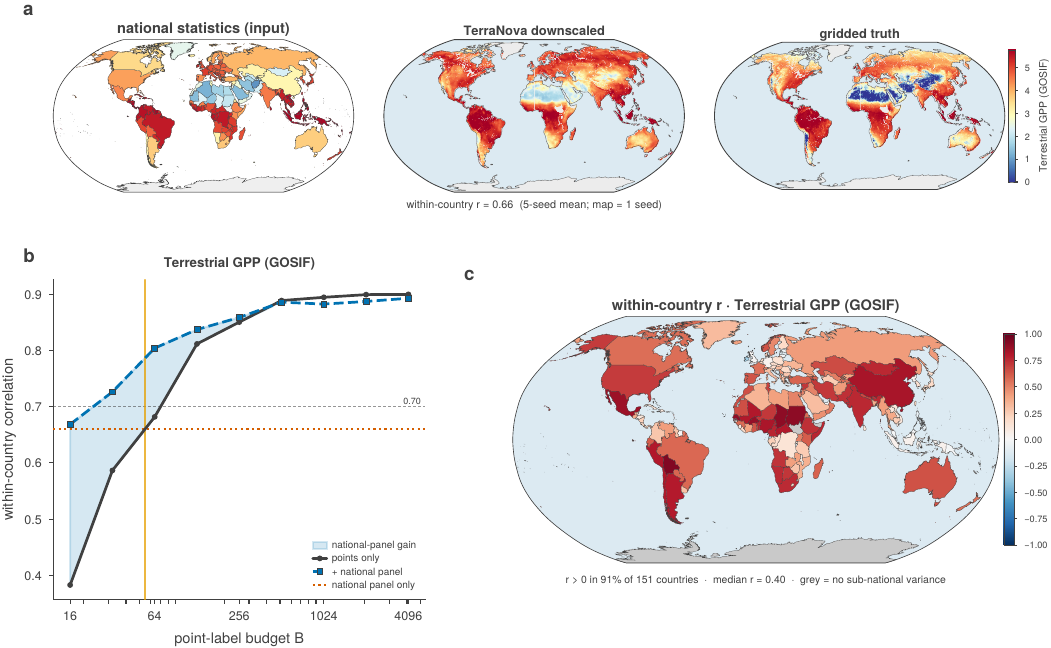}
\caption{\textbf{From national panels to gridded fields.} \textbf{a}, Terrestrial GPP downscaled from national values: the national input, TerraNova's gridded reconstruction and the gridded truth. \textbf{b}, National panels against gridded point labels at matched budgets. \textbf{c}, Within-country correlation per country; grey marks countries with no sub-national variance.}
\label{fig:downscale}
\vspace{-6mm}
\end{figure}

The coupling between each country and the field over its territory can be read in the other direction: a quantity known only at the national level can be downscaled into a gridded field. Adapting the same MiSS operator, TerraNova recovers sub-national structure on four biosphere targets (GPP, fAPAR, LAI and tree cover) at seed-mean within-country correlations of \downQuartet{} against a label-shuffle null of at most \downNullMax{} (Fig.~\ref{fig:downscale}a). It is not carried by a few large countries: across the four targets the correlation is positive in \downPositiveShare{} of the \downCountries{} countries scored, with medians of \downMedianWithin{} (Fig.~\ref{fig:downscale}c). Measured in labels, the national panel is worth about \downExchange{} gridded point labels for this target, and its advantage closes once a few hundred points are available (Fig.~\ref{fig:downscale}b). We report this as the spatial disaggregation the coupled representation makes possible, not as a validated product. Note that this sort of downscaling is not always meaningful (e.g. governance variables may not be downscaled in a intelligible way).

\subsection{Computational cost}\label{sec:cost}

We now report the computational cost of training the model once and of using it afterwards. Adaptation and inference are measured on a consumer laptop, with a mobile \laptopGpu{} with $8$~GB of memory, and on the same machine's CPU. Wall-clock is a median over three seeds and skill a three-seed mean with its standard deviation; energy is integrated from device power sampling with the idle baseline subtracted.

\paragraph{Training.} One production run of $\trainEpochs{}$ epochs reached $\trainSteps{}$~million optimiser steps and $\trainSamples{}$~billion training samples on a single \trainGpuModel{}, at $\trainStepsPerSec{}$ steps per second end to end. Training takes about $\trainHours{}$~GPU-hours, $\trainKWh{}$~kWh and $\trainCarbon{}$~kgCO\textsubscript{2}e, estimated using CodeCarbon~\cite{benoit_courty_2024_11171501}.

\begin{table}[t]
\centering\small
\caption{Every read-out on the frozen backbone at a $300$-label budget on one laptop GPU. The shaded row is the method used throughout the paper and the last row repeats it on the machine's CPU; energy is omitted where it cannot be measured.}
\label{tab:cost}
\begin{tabular}{lrrrcrc}
\toprule
& & \multicolumn{3}{c}{gridded field} & \multicolumn{2}{c}{national indicator} \\
\cmidrule(lr){3-5}\cmidrule(lr){6-7}
Read-out & Trainable & Fit (s) & Energy (kJ) & Skill & Fit (s) & Skill \\
\midrule
Linear probe            & $256$          & $0.3$  & n/a & $0.68 \pm 0.03$ & $<0.1$ & $0.46 \pm 0.12$ \\
Task row only           & $256$          & $50.2$ & $3.1$ & $0.83 \pm 0.02$ & $46.2$ & $0.15 \pm 0.12$ \\
VeRA~\cite{kopiczko2024vera}                   & $101{,}324$    & $55.6$ & $3.2$ & $0.90 \pm 0.05$ & $49.6$ & $0.46 \pm 0.24$ \\
\rowcolor{winrow}  MiSS~\cite{kang2024miss} & $416{,}000$ & $53.3$ & $3.3$ & $0.95 \pm 0.00$ & $49.5$ & $0.60 \pm 0.12$ \\
LoRA~\cite{hu2022lora}                & $803{,}072$    & $53.4$ & $3.3$ & $0.94 \pm 0.02$ & $49.2$ & $0.62 \pm 0.14$ \\
Final-block fine-tuning & $25{,}717{,}508$ & $56.6$ & $3.3$ & $0.96 \pm 0.00$ & $52.0$ & $0.66 \pm 0.12$ \\
\midrule
\textit{MiSS on CPU}    & $416{,}000$    & $599.7$ & n/a & $0.95 \pm 0.01$ & $538.1$ & $0.60 \pm 0.18$ \\
\bottomrule
\end{tabular}
\end{table}

\begin{table}[t]
\centering\small\setlength{\tabcolsep}{5.5pt}
\caption{\textbf{(a)}~Fit wall-clock on the gridded route against budget $K$; the shaded row is the operator used throughout the paper and the italic row repeats it on CPU. \textbf{(b)}~Inference throughput against batch size $B$. A single query is omitted, since it measures dispatch latency rather than the model.}
\label{tab:cost-scale}
\begin{tabular}{lrrrrr}
\toprule
\multicolumn{6}{l}{\textbf{(a)}~Fit wall-clock (s), gridded route} \\
\midrule
Read-out & $K{=}30$ & $100$ & $300$ & $1{,}000$ & $3{,}000$ \\
\cmidrule(lr){2-6}
Linear probe            & $0.3$   & $0.3$   & $0.3$   & $0.3$       & $0.4$ \\
Task row only           & $14.2$  & $22.7$  & $50.2$  & $147.3$     & $148.3$ \\
VeRA~\cite{kopiczko2024vera}                     & $17.9$  & $25.4$  & $55.6$  & $163.8$     & $174.3$ \\
\rowcolor{winrow}  MiSS~\cite{kang2024miss} & $16.9$ & $24.7$ & $53.3$ & $161.9$ & $163.6$ \\
LoRA~\cite{hu2022lora}                    & $16.4$  & $24.4$  & $53.4$  & $157.3$     & $158.2$ \\
Final-block fine-tuning & $20.2$  & $28.2$  & $56.6$  & $158.1$     & $157.3$ \\
\addlinespace
\textit{MiSS on CPU}    & $104.4$ & $246.8$ & $599.7$ & $1{,}428.8$ & $1{,}460.2$ \\
\midrule
\multicolumn{6}{l}{\textbf{(b)}~Inference throughput (queries s$^{-1}$)} \\
\midrule
Route and device & $B{=}8$   & $64$      & $512$      & $4{,}096$  & $16{,}384$ \\
\cmidrule(lr){2-6}
Coordinate, GPU  & $856$     & $5{,}034$ & $9{,}770$  & $10{,}000$ & $9{,}092$ \\
Country, GPU     & $1{,}199$ & $6{,}153$ & $10{,}708$ & $10{,}229$ & $10{,}152$ \\
Coordinate, CPU  & $150$     & $484$     & $732$      & $778$      & $765$ \\
Country, CPU     & $139$     & $396$     & $625$      & $868$      & $875$ \\
\bottomrule
\end{tabular}
\end{table}

\paragraph{Adaptation.} Attaching a new variable takes seconds to minutes on a laptop (Tables~\ref{tab:cost} and~\ref{tab:cost-scale}a). At a given label budget every gradient read-out costs the same to within a few seconds, although they span $256$ to $25.7$~million trainable parameters: the backward pass through the frozen backbone dominates. On gridded fields the linear probe reaches $0.68$ against $0.95$ for the adapted decoder, so most of the added skill costs almost nothing. On national indicators the ordering differs: the probe reaches $0.46$ and beats optimising the task row alone, which never exceeds $0.25$ at any budget, and only the weight-space adapters clear it, at $0.60$ to $0.66$. The same fit on the CPU costs $6.2$ to $11.2$ times the wall-clock depending on budget and geometry, $11.2$ times on the gridded task at this budget, for skill inside the seed spread of the GPU fit. MiSS is the option used throughout the paper because of its efficiency: it matches LoRA on both geometries with half the trainable parameters and a tighter seed spread on the gridded route, and gives up $0.005$ and $0.056$ against final-block fine-tuning while training $1.6\%$ as many parameters.

\paragraph{Inference.} A forward pass costs $\flopsCoord{}$~MFLOP for a coordinate query and $\flopsCountry{}$~MFLOP for a country query, of which the task--spatiotemporal fusion accounts for $\flopsFusionShare{}$ and the temporal fusion a further quarter, against $\modelParams{}$~million parameters in the model. Throughput saturates on the GPU by a batch of $512$ at about $10{,}000$ queries per second, and on the CPU by a batch of $4{,}096$ at about $800$, a twelve-fold gap (Table~\ref{tab:cost-scale}b). Single-query latency is dominated by dispatch rather than by the model, at $9.6$~ms against $102$~$\mu$s per query inside a batch of $512$. The largest batch tested, $16{,}384$ queries, peaked at $3.1$~GiB of allocated tensor memory.

\section{Discussion}
\label{sec:discussion}

Training one model on \numTasks{} variables in two geometries gives the representation properties that neither a gridded geophysical model nor a country-level panel has on its own.

The sparse-reconstruction result is the sharpest of them. A model that has seen a thousand variables over the whole planet \textit{knows} what a plausible field looks like before it sees a single measurement of a new one, so a handful of observations is enough to place that field in a structure the model already holds. The same prior is what makes the latent space readable rather than a lookup table: geography, biome structure and coherent ocean basins are legible in the location embedding, and a development gradient emerges from what the model predicts about each country once the development indicators themselves are removed from the projection. Fields, records and image-derived embeddings admit a compatible representation, although here that compatibility is built by explicit alignment rather than emerging from data alone. The coupling buys capabilities that neither geometry expresses alone, and cross-geometry retrieval and national-to-gridded downscaling are one coupling read in two directions: retrieval names the country a territory belongs to, downscaling writes a national quantity back onto the grid, both supported by the same population-weighted alignment. The ablation shows that the two alignment objectives do separable work, one building the common space and the other enriching it.

Uncertainty is a first-class output. The width responds to evidence, widening as a measurement lattice thins and as a query moves beyond the years a record covers, so a model that reports where it is uncertain flags its own extrapolation instead of failing silently. Cheap adaptation follows from the same architectural choice that produces the benchmark margin: because the decoder is generated per query from the task and the spatiotemporal state, a new variable is reached inside that fusion without touching any pretrained task.

\paragraph{Limitations.} The comparison against specialised geospatial encoders is not capacity-matched (\S\ref{sec:benchmark}). The lead is delivered by the task-conditioned decoder and the fusion it adapts rather than by the frozen embedding, and at the smallest label budgets the strongest baseline is still ahead. Downscaling is a capability of the coupled representation rather than a validated product, and it recovers nothing where a quantity barely varies within countries (\S\ref{sec:downscale}). The socioeconomic and institutional inputs are observational and carry historical and measurement bias that a learned representation can propagate; predictions for under-observed countries carry the widest intervals and should be read with them attached. TerraNova is a representation layer, not a causal model of climate damages, an integrated-assessment model, or a replacement for process-based simulation. It does not identify causal effects, simulate climate-society feedbacks or project policy scenarios, and reading its predictions as if it did would be a misuse. What it provides is a joint observational feature layer in which transfer, spatial inference with uncertainty, cross-domain prediction and hypothesis generation can be studied in one latent space, alongside the causal and mechanistic tools that ultimately answer climate-society questions.

\paragraph{Outlook.} The model treats time as an additional input variable, which limits its ability to learn conditioned dynamics, such as the evolution of an economy under different radiative-forcing assumptions. Lifting TerraNova into a representation better suited to those dynamics could enable high-resolution, uncertainty-aware scenario generation. Beyond the three modalities used here, LLM-derived embeddings of locations, cultures or events could carry information that is not directly quantifiable, although this should be done with the utmost care over bias, transparency, and representation of the Global South. Finally, TerraNova implicitly assumes that countries are somewhat independent, since we do not explicitly model bilateral trade or migration networks; graph neural networks over migration and trade networks, which predict regional economic outcomes at the sub-national scale~\citep{xu2020attentional}, could extend it to cascading climate and economic impacts between countries. These extensions sit on the same foundation this paper sets out: making the coupled data streams of the Anthropocene learnable, transferable and queryable within one model, cheaply enough to put it within reach of groups far from the one that trained it. That is the integrated, cross-disciplinary modelling named a central goal for artificial intelligence in climate research~\citep{ou2026ai}: not a coupling of separate models, but a shared representation in which local environmental structure and national outcomes inform one another directly.

\section*{Broader Impact Statement}

TerraNova lowers the cost of using a planetary-scale model. A new variable is fitted with a compact adapter on a frozen backbone, so a group with a regional survey or a bespoke indicator can specialise the representation on ordinary hardware instead of depending on a supercomputer. This matters most in the data-poor settings where such models are hardest to train from scratch. Calibrated uncertainties are the other half of that story: a model that reports where it is uncertain is harder to trust blindly than a bare point predictor. The limit of this argument is that cheap adaptation distributes the product of a planetary representation while leaving the means of producing one, the corpus, the compute and the labour that assembles both, concentrated where they already were~\cite{mohamed2020decolonial,crawford2021atlas}.

The risks specific to this model follow from its breadth. Because TerraNova learns across such a wide variety of tasks, its representation reaches well beyond physical fields into human development, governance and institutions, and it may learn associations between environmental conditions and national outcomes in a world whose present distribution of those outcomes was produced by history: colonisation, imperialism, extractive economic practices, wars, occupations, trade structure and policy decisions. The model observes none of that history explicitly and does not represent it directly. Its representation therefore recovers real regularities whose causes lie outside its inputs, and the development axis is a statistical description of that confounded world rather than an explanation of it. Reading a learned environment-development association as though geography produced development would recast institutional and historical processes~\cite{acemoglu2001colonial} as natural determinism: a history of relations between places would appear instead as a property of the places themselves, with the relations that produced it no longer visible in the result~\cite{winner1980artifacts}. We report the axis as an insight about the representation, not as a finding about why nations differ.

The corpus carries a viewpoint of its own. National indicators are constructed rather than observed, the product of statistical capacity that is itself unequally distributed, and governance and institutional quality are the hardest of them to pin down. The construct has to be defined before it can be scored, that definition carries a normative model of what institutions ought to look like, and the resulting indices are assembled largely by institutions of the Global North. Rendering such heterogeneous records commensurable, as our per-task standardisation does, is a convenience for training and also a substantive operation: quantities produced under different conditions, by different institutions, for different purposes, are placed on one scale and made available to a single objective~\cite{espeland1998commensuration}. Coverage is unequal for historical reasons, so the model's ignorance is not randomly distributed: it is greatest where its predictions are most likely to be sought. This caution extends across every societal variable the model spans.

The same caution should apply to the model's outputs. A downscaled field is a prediction carrying the visual authority of a measurement, and a caveat in a manuscript does not travel with a raster. Downscaling recovers structure only to the extent that a quantity varies within countries at all, and it fails quietly otherwise, painting a plausible map for a target it cannot resolve. Predictions are least constrained where observations are sparsest, which is also where they are most likely to be taken as the only estimate available. Such surfaces are attractive precisely because they render populations legible to administration at a resolution at which they were never measured~\cite{scott1998seeing}, and that legibility is not an end in itself but an input to distribution: climate finance and loss-and-damage eligibility, humanitarian targeting, subnational resource allocation. That asymmetry is why per-cell uncertainty is part of the prediction here rather than a diagnostic beside it, and why gridded socioeconomic outputs from this model should not be used directly for allocation, targeting or eligibility decisions about the places they describe. Where such uses are contemplated we advise domain expertise and co-design with policy-makers and affected communities, while noting that participation is not a design fix and becomes participation-washing where it is short-term or extractive~\cite{sloane2022participation}. Cheap adaptation can also be problematic. The backbone accepts any new task, so the mechanism that lets, for instance, a public-health group attach a regional indicator also lets anyone attach a target we neither anticipated nor evaluated, inheriting the representation's biases without inheriting its evaluation. Upon acceptance, we will release the model weights and configuration, evaluation code and tutorials under permissive licenses.

\acks{Carlos Rodriguez-Pardo and Massimo Tavoni acknowledge support from the
European Research Council, ERC grant agreement number 101044703 (EUNICE),
CUP D87G22000340006.}

\bibliography{references}

\end{document}